\documentclass[11pt]{article}

\newif\ifpreprint
\preprinttrue

\usepackage[margin=1in]{geometry}
\usepackage[T1]{fontenc}
\usepackage{lmodern}
\usepackage{microtype}
\usepackage{amsmath,amssymb,amsthm}
\usepackage{booktabs}
\usepackage{graphicx}
\usepackage{xcolor}
\usepackage{pgfplots}
\pgfplotsset{compat=1.17}
\usepackage[round,authoryear]{natbib}
\usepackage[colorlinks=true,linkcolor=blue!50!black,citecolor=blue!50!black,urlcolor=blue!50!black]{hyperref}
\usepackage[capitalise,noabbrev]{cleveref}
\usepackage{enumitem}

\newtheorem{theorem}{Theorem}
\newtheorem{lemma}{Lemma}
\newtheorem{corollary}{Corollary}
\newtheorem{proposition}{Proposition}
\theoremstyle{definition}
\newtheorem{definition}{Definition}

\newcommand{\icu}{ICU-Sepsis}
\newcommand{\M}{\mathcal{M}}
\newcommand{\Sset}{\mathcal{S}}
\newcommand{\Tset}{\mathcal{T}}
\newcommand{\Aset}{\mathcal{A}}
\newcommand{\Jterm}{J_{\mathrm{term}}}
\newcommand{\JW}{J_{W}}
\newcommand{\Pterm}{P^{\mathrm{term}}}
\newcommand{\rterm}{r_{\mathrm{term}}}
\newcommand{\vmin}{v_{\min}}
\newcommand{\vmax}{v_{\max}}
\newcommand{\Rmax}{R_{\max}}
\newcommand{\pibhat}{\hat{\pi}_{\beta}}

\definecolor{termc}{HTML}{E8693A}
\definecolor{meanc}{HTML}{2A7AD5}

\title{Which Comparisons Can a Clinical Offline RL Benchmark Support?\\
Interval Reporting for Unsupported Actions}

\ifpreprint
\author{Md.\ Ashraful Kabir Alif \qquad Ashfiqun Mustari \qquad Sadia Afrin\\[4pt]
\normalsize BRAC University, Dhaka, Bangladesh\\[2pt]
\small Contact: \href{mailto:md.ashraful.kabir.alif@g.bracu.ac.bd}{\texttt{md.ashraful.kabir.alif@g.bracu.ac.bd}}}
\else
\author{Anonymous authors\\ \small Paper under double-blind review}
\fi
\date{}

\begin{document}
\maketitle

\begin{abstract}
Offline reinforcement learning benchmarks built from clinical records cannot estimate what happens after treatments
that were rarely given, so they must invent those dynamics. \icu{}, a widely used example, averages the transitions of
well-logged actions (the mean rule); its package also offers sending the patient to death (the terminate rule). The
choice changes conclusions: a uniformly random policy is worth 0.78 under the mean rule and 0.01 under termination.
We show that the two rules bracket the truth. Termination gives a lower bound on a policy's value under every way of
completing the missing dynamics, and the mean rule gives an upper bound when rarely logged actions carry no hidden
benefit. One policy is then certified to beat another when its lower bound exceeds the other's upper bound, and a
one-parameter sensitivity analysis says how much hidden benefit would overturn the claim. On a sepsis simulator with
exactly known dynamics and exact logged rows, none of the 152,237 certified comparisons was wrong; with estimated rows,
most remaining mistakes come from estimating the logged rows. On \icu{}, only 6.6\% of comparisons that involve
rarely logged actions are certified without the no-hidden-benefit assumption. Benchmarks of this kind should report
an interval for each policy.
\end{abstract}

\section{Introduction}\label{sec:intro}

Reinforcement learning (RL) from logged clinical decisions promises treatment policies that improve on routine care
\citep{komorowski2018aiclinician,raghu2017sepsis}, but evaluating such policies offline is fragile
\citep{gottesman2019guidelines,tang2021modelselection,luo2024position}. Benchmarks built from medical records offer a
way out. \icu{} \citep{choudhary2024icusepsis} turns records from the MIMIC-III database \citep{johnson2016mimic} into a
fixed tabular Markov decision process (MDP), so that any policy can be evaluated exactly and compared with any other.

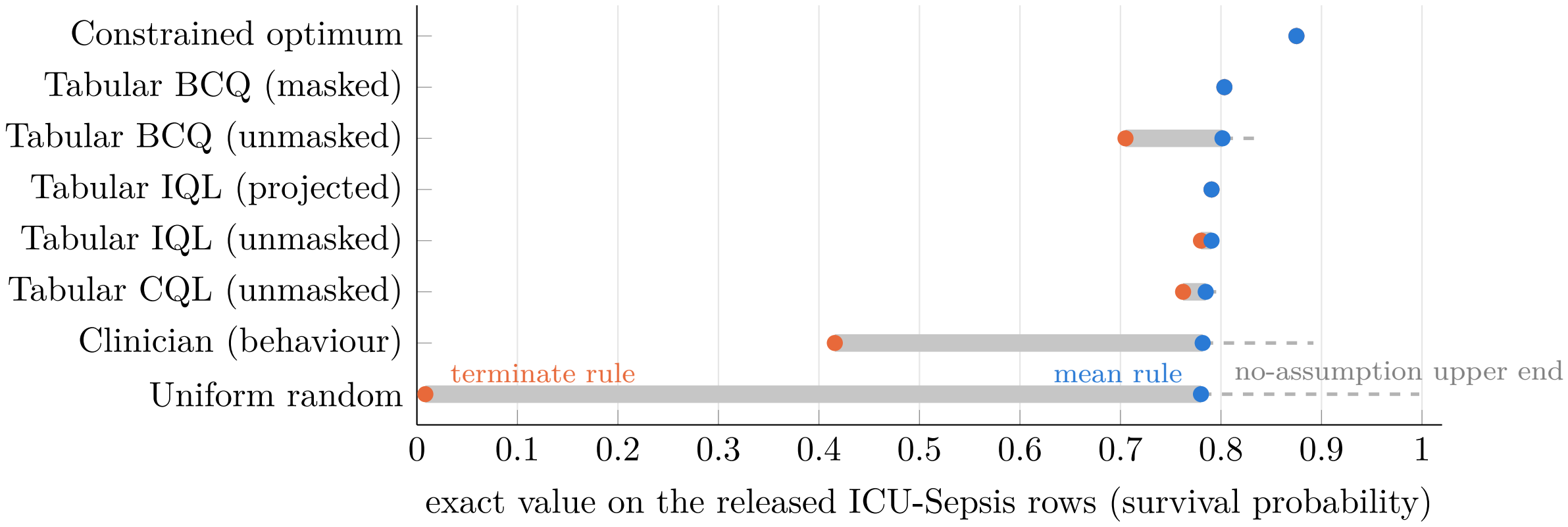
\begin{figure}[t]
\centering
\begin{tikzpicture}
\begin{axis}[
  width=0.72\linewidth, height=5.8cm,
  xmin=0, xmax=1.02, ymin=0.4, ymax=8.6,
  ytick={1,...,8},
  yticklabels={Uniform random, Clinician (behaviour), Tabular CQL (unmasked), Tabular IQL (unmasked),
               Tabular IQL (projected), Tabular BCQ (unmasked), Tabular BCQ (masked), Constrained optimum},
  yticklabel style={font=\small},
  xlabel={exact value on the released \icu{} rows (survival probability)},
  xlabel style={font=\small}, xticklabel style={font=\small},
  xmajorgrids, grid style={gray!20},
  axis lines*=left, clip=false,
]
\foreach \y/\lo/\hi/\up in {1/0.0084/0.7801/0.9973, 2/0.4158/0.7818/0.8920, 3/0.7623/0.7848/0.7980,
                            4/0.7801/0.7906/0.7950, 5/0.7906/0.7906/0.7906, 6/0.7050/0.8015/0.8382,
                            7/0.8034/0.8034/0.8034, 8/0.8751/0.8751/0.8751}{
  \edef\temp{\noexpand\draw[gray!60, line width=1pt, dashed] (axis cs:\hi,\y) -- (axis cs:\up,\y);
             \noexpand\draw[gray!45, line width=5pt] (axis cs:\lo,\y) -- (axis cs:\hi,\y);
             \noexpand\fill[termc] (axis cs:\lo,\y) circle (2.3pt);
             \noexpand\fill[meanc] (axis cs:\hi,\y) circle (2.3pt);}
  \temp
}
\node[font=\scriptsize, anchor=west, text=termc] at (axis cs:0.02,1.42) {terminate rule};
\node[font=\scriptsize, anchor=east, text=meanc] at (axis cs:0.775,1.42) {mean rule};
\node[font=\scriptsize, anchor=west, text=gray] at (axis cs:0.80,1.42) {no-assumption upper end};
\end{axis}
\end{tikzpicture}
\caption{The completion rule decides how good a policy looks on \icu{}. Orange: value under the terminate rule (a lower
bound under every completion). Blue: value under the mean rule (an upper bound if unsupported actions carry no hidden
benefit). Dashed grey: the upper end without that assumption. Policies that never take unsupported actions have a
single value. Numbers are 10-seed means for 8 of the 19 fixed policies of \cref{tab:icu}.}
\label{fig:icu}
\end{figure}

Such a benchmark can only estimate what it has seen. \icu{} estimates the transition distribution of a state--action
pair (a \emph{row}) only when the action was logged more than 20 times in that state. Every other pair is
\emph{unsupported}, and its row has to be invented. By default \icu{} uses the \emph{mean rule}: an unsupported action
behaves like the average of the state's supported actions. Its package also offers a \emph{terminate rule} that sends
the patient to the death state. The benchmark's authors regard unsupported actions as inadmissible and reject death
transitions as artificial \citep[App.~B]{choudhary2024icusepsis}. Learned policies take unsupported actions
nonetheless, and constraint-aware offline RL methods are evaluated on exactly this kind of benchmark
\citep{fujimoto2019bcq,kumar2020cql,laroche2019spibb,le2019batch,zheng2024fisor}. The two rules then disagree, sometimes
completely. On \icu{}, a uniformly random policy is worth 0.780 under the mean rule, about as much as the clinicians
(0.782), but only 0.008 under termination; the clinician policy itself drops to 0.416 (\cref{fig:icu}).

Which number is right? The data cannot say, so we ask a different question: what do the supported rows imply about a
policy's value under \emph{every} way of completing the unsupported ones? The answer is an interval with the two rules
at its ends. If death is the worst possible outcome, the terminate rule is a lower bound under every completion. The
mean rule, like any convex average of supported rows, is an upper bound if unsupported actions carry \emph{no hidden
benefit} (NHB): in truth they are no better than the supported actions the average stands in for. A comparison between
two policies is \emph{certified} when the first policy's lower bound exceeds the second policy's upper bound. Such a
claim does not depend on which completion rule the benchmark uses. If the claim needs NHB, a one-parameter sensitivity
analysis reports how much hidden benefit would overturn it.

We test this protocol where the truth is known. We rebuild the \icu{} construction on an exactly enumerated
Gumbel-max sepsis simulator \citep{oberst2019gumbel} and on Garnet MDPs \citep{archibald1995garnet}, evaluate policies
from 20 tabular offline RL methods exactly, and compare every certified claim with the true ranking. We then apply the
protocol to \icu{} itself, which needs no ground truth because the bounds hold for any completion of its released rows.
We propose no new learning method. Our contributions are:
\begin{itemize}[leftmargin=1.4em, itemsep=2pt, topsep=3pt]
\item \textbf{An identification result for completed benchmarks} (\cref{sec:theory}). Under NHB, the terminate rule
and any convex imputation are the two ends of the exact set of values that the supported rows allow
(\cref{thm:bracket}). This yields a certificate for pairwise comparisons, an assumption-free version of it, and a
sensitivity family between the two with a breakdown value per claim.
\item \textbf{A validation against exact ground truth} (\cref{sec:validation}). With exact supported rows, certified
claims on the sepsis simulator were never wrong, including the 41,907 claims that rest on NHB; on Garnet MDPs, errors
appear where NHB fails by design, and the breakdown value flags them. With estimated rows, about one certified claim in
ten is wrong, and we trace these errors to estimation and state aggregation rather than to the completion rule
(\cref{sec:limits}).
\item \textbf{An analysis of \icu{}} (\cref{sec:icu}). Only 6.6\% of the comparisons in which the completion matters
are certified without NHB, and half of the claims that need NHB break at 9\% of the largest possible hidden benefit. We
turn these findings into a four-step reporting protocol (\cref{sec:recommendations}).
\end{itemize}

\section{Setting: benchmarks built from records}\label{sec:setting}

\paragraph{Absorbing MDPs and exact evaluation.}
An environment is a finite absorbing MDP $(\Sset\cup\Tset,\Aset,P,R,d_0)$ with live states $\Sset$, absorbing states
$\Tset$ (for example death, discharge or timeout) and initial distribution $d_0$ on $\Sset$. A reward $R(s)$ is paid on
entering $s\in\Tset$, live states pay nothing, and there is no discounting. Every value therefore lies in
$[\vmin,\vmax]$ with $\vmin=\min(0,\min_\Tset R)$ and $\vmax=\max(0,\max_\Tset R)$; we write $\Rmax=\max_\Tset R$. A
stationary policy $\pi$ is \emph{proper} under rows $P$ if it is absorbed with probability one from every live state.
Its value $v_P^\pi$ then solves one linear system, which gives the return $J_P(\pi)=d_0^\top v_P^\pi$ and the occupancy
$\mu_P^\pi(s)$, the expected number of visits to $s$. We write $U_P^\pi=R+v_P^\pi$ (with $v=0$ on $\Tset$) for the value
of arriving in a state. All policies in this paper are evaluated only under rows for which they are proper; this holds
in all our environments (\cref{app:env}).

\paragraph{Support and completion rules.}
Each live state $s$ has a set $\M(s)\subseteq\Aset$ of supported actions, those logged often enough. A policy is
\emph{admissible} if it never takes an unsupported action. Let $p_\pi$ be the probability that $\pi$ takes at least one
unsupported action before absorption. Histories up to that first decision use only supported rows, so $p_\pi$ is the
same under every completion. A benchmark must fill in the rows of unsupported pairs. A \emph{convex imputation} with
weights $W(\cdot\mid s,a)$, a distribution over $\M(s)$, sets
$P_W(\cdot\mid s,a)=\sum_{a'\in\M(s)}W(a'\mid s,a)\,P(\cdot\mid s,a')$ for $a\notin\M(s)$; the mean rule takes $W$ uniform.
The terminate rule sends every unsupported action to a sink state $s^\dagger$ with reward $\rterm=R(s^\dagger)$. We
write $\JW(\pi)$ (or $J_{\mathrm{mean}}$ for the mean rule) and $\Jterm(\pi)$ for the two resulting values. In all our
environments the sink is a worst absorbing state, $\rterm=\vmin$: death pays 0 on \icu{} and Garnet and $-1$ on the
Gumbel simulator.

\paragraph{The construction we study.}
\icu{} follows a recipe that we can apply to any simulator \citep{choudhary2024icusepsis}. Given a true system, a
physician policy, $n$ hospital episodes and a count threshold $\tau$, we (i) simulate the episodes and count visits;
(ii) mark $a$ supported in $s$ if it was logged at least $\tau$ times (\icu{}'s ``more than 20'' is $\tau=21$);
(iii) remove states with no supported action; (iv) estimate supported rows by empirical frequencies; (v) complete the
rest by the mean rule, with termination available at evaluation; and (vi) estimate the behaviour policy, from which
user datasets for offline learners are simulated on the benchmark. Full details are in \cref{app:env}.

\section{What the supported rows identify}\label{sec:theory}

Fix one benchmark: its states, rewards, initial distribution, support sets and supported rows. A \emph{completion} is
any set of rows that keeps all of these and assigns some distribution to each unsupported pair. $P_W$ and $\Pterm$ are
completions, and so is the unknown truth $P^\star$: on a simulator, the true system restricted to the benchmark's mask;
on \icu{}, any dynamics that agree with its released supported rows. Everything below rests on one exact identity
(\cref{lem:sim}, a form of the simulation lemma of \citealp{kearns2002e3}): if $\pi$ is proper under two completions
$P_0$ and $P_1$, only unsupported pairs contribute to the difference in value,
\begin{equation}\label{eq:sim}
J_{P_0}(\pi)-J_{P_1}(\pi)=\sum_{s\in\Sset}\mu_{P_1}^\pi(s)\sum_{a\notin\M(s)}\pi(a\mid s)\,
\big(P_0-P_1\big)(\cdot\mid s,a)^\top U_{P_0}^\pi .
\end{equation}

\begin{definition}[No hidden benefit]\label{def:nhb}
Let $W$ be a convex imputation and $\pi$ proper under $P_W$. A completion $P^\star$ has \emph{no hidden benefit} (NHB)
for $\pi$ if every live $s$ and unsupported $a$ with $\pi(a\mid s)>0$ satisfy
$P^\star(\cdot\mid s,a)^\top U_{P_W}^\pi\le P_W(\cdot\mid s,a)^\top U_{P_W}^\pi$. It has \emph{intermediate harm} if
every unsupported row is $P^\star(\cdot\mid s,a)=(1-\delta_{s,a})P_W(\cdot\mid s,a)+\delta_{s,a}\,\mathbb{1}_{s^\dagger}$
for some $\delta_{s,a}\in[0,1]$.
\end{definition}

NHB is a one-step condition: taking an unsupported action once and then following $\pi$ is, in truth, no better than
taking the supported actions that the imputation averages over. Intermediate harm is the clinically readable special
case. A rarely given treatment behaves like the usual ones, except that it carries some extra risk of death. When
$\rterm=\vmin$, intermediate harm implies NHB for every policy. NHB cannot be checked from the released rows, which is
why we pair it with an assumption-free test and a sensitivity analysis below.

\begin{theorem}[Bracketing]\label{thm:bracket}
Let $\rterm=\vmin$, let $W$ be a convex imputation, let $P^\star$ be a completion, and let $\pi$ be proper under
$P^\star$ and $P_W$ (and hence under $\Pterm$).
\begin{enumerate}[label=(\roman*), itemsep=1pt, topsep=2pt]
\item \emph{Lower bound, no assumption.} $\Jterm(\pi)\le J_{P^\star}(\pi)$.
\item \emph{Upper bound under NHB.} If $P^\star$ has NHB for $\pi$, then $J_{P^\star}(\pi)\le \JW(\pi)$, so that
$\Jterm(\pi)\le J_{P^\star}(\pi)\le\JW(\pi)$.
\item \emph{Width.} $0\le\JW(\pi)-\Jterm(\pi)\le p_\pi(\vmax-\vmin)$; if $p_\pi=0$ (for instance, if $\pi$ is
admissible), every completion gives $\pi$ the same value.
\item \emph{Tightness.} Every value in $[\Jterm(\pi),\JW(\pi)]$ is attained by intermediate harm with a constant
$\delta$, so this interval is exactly the set of values of $\pi$ under NHB. Without NHB, the values of $\pi$ over all
completions fill $[\Jterm(\pi),\,\Jterm(\pi)+p_\pi(\Rmax-\vmin)]$.
\end{enumerate}
\end{theorem}

\begin{proof}[Proof sketch]
For (i), apply \eqref{eq:sim} with $P_0=P^\star$ and $P_1=\Pterm$. Each unsupported pair contributes a weight times
$P^\star(\cdot\mid s,a)^\top U_{P^\star}^\pi-\rterm\ge0$, because values are averages of absorbing rewards and so are
at least $\vmin=\rterm$. For (ii), apply \eqref{eq:sim} with $P_0=P_W$ and $P_1=P^\star$; each term is nonnegative by
NHB. For (iii), take $P_0=P_W$ and $P_1=\Pterm$: the weights sum to $p_\pi$ and each term lies in $[0,\vmax-\vmin]$.
Part (iv) follows from continuity in $\delta$ and from the completion that sends unsupported actions to the best
absorbing state. Full proofs are in \cref{app:proofs}.
\end{proof}

\begin{corollary}[Certified comparison]\label{cor:cert}
Let $\rterm=\vmin$ and let $\pi_1,\pi_2$ be proper under $P^\star$ and $P_W$.
If $\Jterm(\pi_1)>\JW(\pi_2)$, then $J_{P^\star}(\pi_1)>J_{P^\star}(\pi_2)$ whenever $P^\star$ has NHB for $\pi_2$; no
assumption is needed about how the unsupported actions of $\pi_1$ behave. If even
$\Jterm(\pi_1)>\Jterm(\pi_2)+p_{\pi_2}(\Rmax-\vmin)$, then $J_{P^\star}(\pi_1)>J_{P^\star}(\pi_2)$ for every completion
(an \emph{NHB-free} certificate). Either way, both single rules rank $\pi_1$ first.
\end{corollary}

\paragraph{How much hidden benefit would overturn a claim?}
Between the two tests lies a family of certificates. Let
$E(\pi)=p_\pi(\Rmax-\vmin)-\big(\JW(\pi)-\Jterm(\pi)\big)\ge0$ be the largest amount by which any completion can lift
$\pi$ above $\JW(\pi)$ (\cref{thm:bracket}(iv)). For $\kappa\in[0,1]$, the test
\begin{equation}\label{eq:kappa}
\Jterm(\pi_1)>\JW(\pi_2)+\kappa\,E(\pi_2)
\end{equation}
certifies $\pi_1$ over $\pi_2$ under every completion that lifts $\pi_2$ by at most a fraction $\kappa$ of that largest
hidden benefit. $\kappa=0$ is the NHB test and $\kappa=1$ the NHB-free one. The largest $\kappa$ a claim passes, its
\emph{breakdown value} $\kappa^*$, measures how robust the claim is to hidden benefit.

\paragraph{Why constraint-aware methods gain nothing on a completed benchmark.}
The bracket also explains a pattern that otherwise looks puzzling. Under a convex imputation, every policy has an
admissible twin with exactly the same value: move each unsupported action's probability onto the supported actions
with the imputation's weights. Under termination with $\rterm=\vmin$, an unsupported action is never better than a
supported one. In both cases the best admissible policy is optimal among all policies, so restricting a learner to
supported actions costs nothing on the benchmark, whatever unsupported actions are worth in truth
(\cref{prop:flat} in \cref{app:proofs}; the mean-rule half is noted by \citealp{choudhary2024icusepsis}). Rules can
therefore reorder only policies that leave the support, and those are exactly the policies whose comparisons
\cref{cor:cert} examines.

\section{Which comparisons survive on \icu{}}\label{sec:icu}

\icu{} (version 2) has 713 live states and 25 actions. Its released tensor supports 2,238 live state--action pairs,
and each of the other 15,587 rows equals the mean of its state's supported rows to within $10^{-12}$; survival pays 1
and death 0, so the death state is a worst absorbing state. No ground truth exists, but none is needed:
\cref{thm:bracket} bounds every completion of the released supported rows. We analyse two sets of policies. The first
consists of the 50 default policies that our 20 tabular methods (\cref{sec:validation}) produce on each of 50 user
datasets of 10,000 simulated clinician episodes. The second consists of 19 policies fixed before this analysis, among
them tabular BCQ, IQL and CQL in several extraction modes, neural and diffusion policies, the clinician and a uniformly
random policy (\cref{app:methods}). \cref{tab:icu} and \cref{fig:icu} show intervals for representative members of the
second set.

\begin{table}[t]
\centering
\caption{Intervals on the released \icu{} rows (10-seed means) for 8 of the 19 fixed policies. NHB-free upper: the upper
end without any assumption. Certified above (below): of the other 18 policies, how many this policy is certified to
beat (lose to); NHB-free counts in parentheses. Rank: position under the mean rule.}
\label{tab:icu}
\small
\setlength{\tabcolsep}{4pt}
\begin{tabular}{lrrrrrrrr}
\toprule
 & & & & & NHB-free & \multicolumn{2}{c}{certified} & mean \\
\cmidrule(lr){7-8}
Policy & $\Jterm$ & $J_{\mathrm{mean}}$ & width & $p_\pi$ & upper & above & below & rank \\
\midrule
Constrained optimum & 0.8751 & 0.8751 & 0.0000 & 0.000 & 0.8751 & 18 (16) & 0 (0) & 1 \\
Tabular BCQ, masked & 0.8034 & 0.8034 & 0.0000 & 0.000 & 0.8034 & 17 (14) & 1 (1) & 2 \\
Tabular BCQ, unmasked & 0.7050 & 0.8015 & 0.0966 & 0.133 & 0.8382 & 0 (0) & 2 (1) & 3 \\
Tabular IQL, projected & 0.7906 & 0.7906 & 0.0000 & 0.000 & 0.7906 & 13 (10) & 3 (3) & 5 \\
Tabular IQL, unmasked & 0.7801 & 0.7906 & 0.0105 & 0.015 & 0.7950 & 4 (3) & 3 (2) & 5 \\
Tabular CQL, unmasked & 0.7623 & 0.7848 & 0.0225 & 0.036 & 0.7980 & 0 (0) & 7 (2) & 10 \\
Clinician (behaviour) & 0.4158 & 0.7818 & 0.3660 & 0.476 & 0.8920 & 0 (0) & 11 (0) & 15 \\
Uniform random & 0.0084 & 0.7801 & 0.7717 & 0.989 & 0.9973 & 0 (0) & 12 (0) & 16 \\
\bottomrule
\end{tabular}
\end{table}

\paragraph{Most comparisons that involve unsupported actions are not identified.}
Across the 50 datasets, 72.9\% of pairs of default policies are certified, and 53.5\% without NHB. These numbers look
reassuring, but almost all certified pairs compare two admissible policies, whose values do not depend on the
completion at all (\cref{thm:bracket}(iii)). Among the pairs that involve at least one policy with $p_\pi>0$, those
where the completion matters, only 6.6\% are certified without NHB (\cref{tab:kappa}). Assuming NHB for the losing
policy raises this to 46.3\%, but the gain is fragile. At a tenth of the largest possible hidden benefit ($\kappa=0.1$)
only 24.9\% of these pairs remain certified, and half of the claims that need NHB break at $\kappa^*\le0.09$. To make
this concrete: for unmasked tabular BCQ ($p_\pi=0.133$) the largest hidden benefit is $E=0.037$, so $\kappa=0.1$ allows
its rarely logged actions to raise survival by at most 0.4 percentage points above the mean-rule value. Claims that
break at that level are settled by the completion rule; the data alone cannot decide them.

\paragraph{What this means for reported rankings.}
The 19 fixed policies tell the same story (\cref{tab:icu}). Of their 80 pairs that involve a policy with $p_\pi>0$, 8
are certified without NHB and 38 with it. The mean rule ranks unmasked tabular BCQ third, yet its interval certifies it
below the constrained optimum without NHB and below masked BCQ only with NHB, and above no other policy. Without NHB,
no policy is certified to beat the clinician or the uniformly random policy: some completion of the released rows
gives them values as high as 0.892 and 0.997, if rarely logged actions were in fact beneficial. A leaderboard built on the mean rule alone would hide
all of this.

\begin{table}[t]
\centering
\caption{Hidden-benefit sensitivity. Share (\%) of completion-sensitive pairs (at least one policy with $p_\pi>0$)
certified by test \eqref{eq:kappa}, from the NHB test ($\kappa=0$) to the NHB-free test ($\kappa=1$). In parentheses
(simulators only): share of these certified claims that the truth reverses or ties. Last column: median breakdown
value $\kappa^*$ of the claims that need NHB. Default policies; the last row counts pairs out of 80.}
\label{tab:kappa}
\small
\setlength{\tabcolsep}{4pt}
\begin{tabular}{lrrrrrr}
\toprule
 & \multicolumn{5}{c}{$\kappa$} & median \\
\cmidrule(lr){2-6}
Setting & 0 (NHB) & 0.1 & 0.25 & 0.5 & 1 (NHB-free) & $\kappa^*$ \\
\midrule
Gumbel, true rows & 62.0 (0) & 46.8 (0) & 33.1 (0) & 15.9 (0) & 10.9 (0) & 0.20 \\
Garnet, true rows & 67.7 (2.9) & 64.4 (2.0) & 60.7 (0.9) & 55.8 ($<$0.1) & 47.5 (0) & 0.39 \\
\quad $h={-}0.5$ (NHB fails) & 55.2 (13.2) & 52.6 (9.0) & 49.8 (4.2) & 47.5 (0.2) & 44.4 (0) & 0.25 \\
Gumbel, estimated rows & 60.8 (2.7) & 49.0 (2.9) & 34.5 (3.2) & 16.2 (4.5) & 10.2 (5.2) & 0.23 \\
Garnet, estimated rows & 70.5 (7.8) & 68.6 (7.0) & 65.8 (6.1) & 61.9 (5.1) & 54.8 (4.0) & 0.45 \\
\icu{}, 50 datasets & 46.3 & 24.9 & 14.6 & 9.5 & 6.6 & 0.09 \\
\icu{}, 19 fixed policies & 38/80 & 15/80 & 13/80 & 9/80 & 8/80 & 0.05 \\
\bottomrule
\end{tabular}
\end{table}

\section{Validation where the truth is known}\label{sec:validation}

The results on \icu{} are only as trustworthy as the bracket itself, so we test it on systems whose dynamics we know.

\paragraph{Environments.}
We apply the construction of \cref{sec:setting} to two simulators. The Gumbel-max sepsis simulator
\citep{oberst2019gumbel} is enumerated exactly from its published code (606 live states, 8 actions), with a per-step
timeout probability of 0.05. Its physician puts probability $1-\epsilon$ on the optimal action. Four cells cross
$n\in\{2{,}000,\,10{,}000\}$ hospital episodes with $\tau\in\{5,20\}$ at $\epsilon=0.05$. Two more, at $n=10{,}000$ and
$\tau=20$, use $\epsilon=0.2$ and a physician who rarely gives vasopressors even though they can help. An oracle cell
keeps the true supported rows. In
Garnet MDPs \citep{archibald1995garnet} (150 live states, 8 actions), the behaviour policy favours the one to three best
actions of each state. A harm parameter $h$ sets the true row of every other action to a mixture of the preferred
actions' mean row and a jump to death ($h>0$) or to the goal ($h<0$). Where the mask contains the preferred actions,
$h\in[0,1]$ is intermediate harm and $h<0$ violates NHB by design. We cross $h\in\{-0.5,0,0.5,1\}$ with
$\epsilon\in\{0.05,0.3\}$. Every cell has 30 independent dataset draws; further cells are described in \cref{app:env}.

\paragraph{Methods and protocol.}
We run 20 exact tabular methods from six families: behaviour baselines; CQL \citep{kumar2020cql} and IQL
\citep{kostrikov2022iql}; discrete BCQ \citep{fujimoto2019benchmarking}, VI-LCB
\citep{rashidinejad2021pessimism,jin2021pessimism}, MBS-QI \citep{liu2020mbsqi} and SPIBB \citep{laroche2019spibb};
Lagrangian model-based, CMDP-LP \citep{altman1999cmdp,le2019batch,lee2022coptidice} and a tabular FISOR
\citep{zheng2024fisor}; SfBC \citep{chen2023sfbc} and IDQL \citep{hansenestruch2023idql}; and exact categorical
diffusion \citep{austin2021d3pm}. Each method yields a native policy, its admissible projection and, for eight methods,
a masked extraction, giving 50 default policies per draw (\cref{app:methods}). Counting every hyper-parameter variant,
each draw has 110 policies: checks of the bounds use all 110, and pairwise comparisons use the 50 default policies. For
every policy we compute two intervals. The \emph{true-rows} interval keeps the true supported rows, so \cref{thm:bracket} applies exactly with the
truth as $P^\star$. The \emph{benchmark} interval uses the estimated rows, as a real benchmark would. A certified claim
is an error if the truth orders the pair differently or ties it. Three hypotheses about single rules (H1--H3,
\cref{app:results}) were fixed in code before the first run; all other analyses are exploratory.

\begin{figure}[t]
\centering
\includegraphics[width=0.92\linewidth]{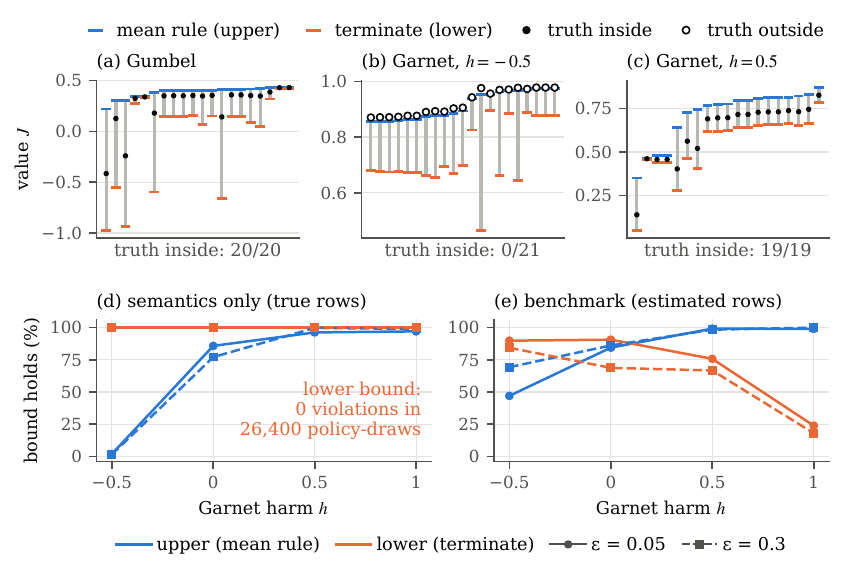}
\caption{The bracket on known systems. (a)--(c) One draw of the Gumbel simulator ($n=10$k, $\tau=20$) and of Garnet:
true value and true-rows interval of each policy with $p_\pi>0$. Where NHB fails by design ($h=-0.5$), the truth lies
above the mean rule. (d) Garnet, 30 draws of all 110 policies: share of policies within each bound with true supported
rows. (e) The same with estimated rows.}
\label{fig:bracket}
\end{figure}

\paragraph{The bracket holds where the theorem says it should.}
With true supported rows, the lower bound held in all 46,200 policy-draws of the 14 estimated cells (110 policies,
30 draws each) and in the oracle cell, as \cref{thm:bracket}(i) guarantees; this mainly checks the code (\cref{fig:bracket}). The upper bound needs NHB.
It held for 99.0\% of policy-draws with $p_\pi>0$ on the Gumbel simulator and for 98.4\% on Garnet with $h\ge0.5$, and
it failed almost always at $h=-0.5$, where hidden benefit is built in.

\paragraph{On the sepsis simulator, certified claims were never wrong.}
On the Gumbel simulator with true rows, 69.0\% of pairs of default policies are certified, and none of the 152,237
certified claims is wrong (\cref{tab:robust}). Most of them (110,330) are NHB-free and cannot be wrong by
\cref{cor:cert}. The real test is the other 41,907 claims, which rest on NHB: all of them are right, including the
11,766 that compare distinct policies. Single rules are much less reliable. On pairs the certificate leaves open, the
mean rule's strict claims are wrong 60.5\% of the time and termination's 15.7\%. The result is not an artefact of
physicians who leave a single supported action in most states. With softer physicians ($\epsilon=0.3$ and $0.5$; one
supported action in 81\% and 58\% of decision states instead of 96\%), none of their 8,382 and 8,284 NHB-dependent
claims is wrong (\cref{tab:sens}). Nor is it abstention alone: a rule that claims the same number of pairs, choosing those with the
largest gaps, errs in 1.6--5.7\% of its claims.

\paragraph{Where NHB fails, the breakdown value flags the failures.}
On Garnet with true rows, 1.0\% of certified claims are wrong, almost all at $h=-0.5$; at $h\ge0.5$ only 10 of
132,422 are. At $h=-0.5$, 2,342 of the 3,459 claims that need NHB are wrong, as they should be when the assumption is
false. The breakdown value separates them: only 3.0\% of the claims with $\kappa^*>0.5$ are wrong. NHB-free claims
remain correct throughout.

\begin{table}[t]
\centering
\caption{Certificates against the truth (\%). Share: share of pairs of default policies certified. Err.: share of
certified claims the truth reverses or ties. Needs NHB: error among claims certified only with NHB. Margin rule: per
draw, a single rule (mean, terminate or the interval midpoint) claims as many pairs as the NHB certificate, choosing
the largest gaps.}
\label{tab:robust}
\footnotesize
\setlength{\tabcolsep}{3pt}
\begin{tabular}{lrrrrrrrr}
\toprule
 & \multicolumn{2}{c}{NHB certificate} & \multicolumn{2}{c}{NHB-free certificate} & needs NHB & \multicolumn{3}{c}{margin rule, same coverage} \\
\cmidrule(lr){2-3}\cmidrule(lr){4-5}\cmidrule(lr){6-6}\cmidrule(lr){7-9}
Setting & share & err. & share & err. & err. & mean & term & mid \\
\midrule
Gumbel, true rows (6 cells) & 69.0 & 0 & 50.0 & 0 & 0 & 5.7 & 2.8 & 1.6 \\
Garnet, true rows (8 cells) & 85.0 & 1.0 & 76.7 & 0 & 9.8 & 2.1 & 9.9 & 8.0 \\
\quad of which $h={-}0.5$ (2 cells) & 79.0 & 4.0 & 74.2 & 0 & 67.7 & 0.8 & 23.5 & 17.7 \\
\midrule
Gumbel, estimated rows & 68.6 & 10.6 & 49.9 & 13.7 & 2.2 & 15.0 & 9.0 & 8.1 \\
Garnet, estimated rows & 86.3 & 9.2 & 79.9 & 8.3 & 21.1 & 7.9 & 12.8 & 11.3 \\
\bottomrule
\end{tabular}
\end{table}

\paragraph{Which single rule is more faithful depends on the sign of the hidden benefit.}
If a benchmark must report one number, the theory suggests which. Under NHB a policy is worth at most its admissible
twin; termination preserves this order, the mean rule reports a tie, and hidden benefit can reverse it. On the Gumbel
simulator, where unsupported actions are mostly harmful, termination ranked policies closer to the truth in all six
estimated cells (pre-specified H1; Kendall $\tau_b$ higher by 0.02 to 0.43, Holm-corrected $p\le0.0037$). On Garnet the
winner flips with the design: the mean rule ranks better for $h\le0$ and termination for $h\ge0.5$. Across all 15
simulator cells, termination won exactly where the aggregate hidden benefit was negative (\cref{app:results}).

\section{What intervals do not fix}\label{sec:limits}

\cref{thm:bracket} takes the supported rows as given. A real benchmark estimates them, and \cref{sec:validation}
shows that this matters more than the completion rule.

\paragraph{Estimation error dominates once rows are estimated.}
With estimated rows, the benchmark interval contains the true value for 76.4\% of Gumbel policy-draws with $p_\pi>0$
(all 110 policies) and for 49.3\% on Garnet (\cref{fig:bracket}e). Certified claims are wrong in 10.6\% of cases on Gumbel (draw-clustered
95\% interval 9.4--11.7\%) and 9.2\% on Garnet (8.8--9.7\%), and a single rule claiming as many pairs does about as
well (\cref{tab:robust}). Most of these errors compare two admissible policies: 14.4\% of such Gumbel claims are wrong,
against 2.7\% of the others. Admissible policies have the same value under every completion, so these are estimation
errors, not completion errors, and no completion analysis can remove them. The oracle cell confirms this: with true
supported rows it keeps 97.7\% coverage and no certified error. An exact decomposition (\cref{app:results}) splits the
gap between benchmark and truth into a completion term, which \cref{thm:bracket} controls, and an estimation term,
which barely depends on the rule and shrinks with more data.

\paragraph{Estimation error can reverse certified rankings.}
In the Gumbel cell with $n=2{,}000$ and $\tau=5$, five admissible methods (masked VI-LCB, projected SPIBB, Lagrangian
model-based, CMDP-LP and guided diffusion) beat projected IQL on the benchmark (means 0.444--0.453 against 0.440) but
lose to it in truth (0.364--0.422 against 0.431). The truth reverses all 89 of their certified wins. Because all five
policies are admissible, the reversals come entirely from estimation, consistent with planning that inherits the
model's optimism.

\paragraph{Valid finite-sample bounds are too wide to decide anything.}
Bootstrap resampling of the hospital episodes widens the intervals and cuts certified errors from 11.0\% to 2.4\%, but
certifies only 49.6\% of pairs, and it is a heuristic. Joint 95\% $L_1$ confidence sets for the rows
\citep{weissman2003l1} give valid intervals (coverage 100\%), but at 2,000--10,000 hospital episodes they span 94\%
(Gumbel) and 100\% (Garnet) of the value range and certify no pair at all (\cref{app:results}). Bounding estimation
error at realistic sample sizes needs structure, such as factored or smoothed rows, and structure is itself an
unchecked modelling assumption.

\paragraph{Aggregating states adds errors that no completion analysis can bound.}
\icu{} estimates its MDP on clustered states. When we estimate the Gumbel benchmark on aggregated observations,
unsupported actions become nearly worthless to the true optimum (at most 0.001), yet 43.7\% (diabetes hidden) and 9.8\%
(diabetes and glucose hidden) of certified claims are wrong, against 7.5\% without aggregation. Hiding a confounder is
a model error that sits outside the completion question.

\section{Related work}\label{sec:related}

\paragraph{Partial identification, robust MDPs and pessimism.}
\cref{thm:bracket} is partial identification in the sense of \citet{manski2003partial}: the benchmark fixes the
supported rows and leaves the others free. The terminate and imputed values are the worst and best cases of a fixed
policy over an $(s,a)$-rectangular uncertainty set, as in robust MDPs
\citep{iyengar2005robust,nilim2005robust,wiesemann2013robust}. Termination at the worst reward is the fixed-policy
counterpart of the absorbing HALT state of MOReL \citep{kidambi2020morel} and of pessimism in offline RL
\citep{jin2021pessimism}; the assumption-free upper end mirrors the optimistic absorbing state of R-MAX
\citep{brafman2002rmax}. Interval dominance over all completions, our NHB-free certificate, follows directly from these
ideas, and the simulation identity \eqref{eq:sim} is classical \citep{kearns2002e3}. What is new is the identified set
under NHB with the two standard completion rules as its exact ends, the certificate and sensitivity family it yields,
and the measurement against exactly known dynamics of how often rules and certificates err and why. Off-policy
evaluation beyond overlap narrows similar bounds with smoothness assumptions
\citep{khan2024overlap,wang2026network}; here the overlap gap is the benchmark's mask, and NHB plays the role of the
smoothness assumption. \citet{kallus2020confounding} bound policy values under unobserved confounding, and Delphic
offline RL \citep{pace2024delphic} separates uncertainty due to non-identifiable confounding and penalises it during
learning. We evaluate rather than learn, and our aggregated cells show that confounding hidden by a benchmark's state
space causes errors that completion bounds do not cover.

\paragraph{Clinical offline RL, simulators and benchmarks.}
\icu{} \citep{choudhary2024icusepsis} turns the AI Clinician pipeline \citep{komorowski2018aiclinician,raghu2017sepsis}
into a benchmark MDP. The Gumbel-max simulator \citep{oberst2019gumbel} has served as known truth for model selection
\citep{tang2021modelselection} and factored action spaces \citep{tang2022factored}. Offline RL methods stay near the
data by filtering, support matching, pessimism or constraints
\citep{levine2020offline,fujimoto2019bcq,kumar2019bear,kumar2020cql,kostrikov2022iql,laroche2019spibb,rashidinejad2021pessimism,le2019batch,tessler2019rcpo,xu2022cpq,lee2022coptidice,zheng2024fisor,liu2024osrl};
all of them take the environment as given. Invalid-action masking \citep{huang2022masking} and shielding
\citep{alshiekh2018shielding} assume a known set of invalid actions. Work on D4RL \citep{fu2020d4rl}, DOPE
\citep{fu2021dope}, off-policy evaluation \citep{voloshin2021ope} and empirical practice
\citep{patterson2024empirical,agarwal2021precipice} shows that conclusions depend on how data, environments and
statistics are handled. To our knowledge, no prior work bounds the effect of the completion rule of an \icu{}-style
benchmark or measures its rank fidelity against exactly known dynamics.

\section{Recommendations and limitations}\label{sec:recommendations}

\paragraph{A reporting protocol.}
For benchmarks that complete unsupported actions, we recommend four steps.
\begin{enumerate}[leftmargin=1.6em, itemsep=1pt, topsep=2pt]
\item Report $[\Jterm(\pi),\JW(\pi)]$, $p_\pi$ and the NHB-free upper end for every policy, with the sink a worst
absorbing state.
\item Present NHB-free comparisons as robust to the completion; they still say nothing direct about patients. Report
comparisons that need NHB separately, with their breakdown value $\kappa^*$, and flag the rest as decided by the completion rule.
\item Release both completion rules, the support mask and its visit counts, where data-use agreements allow.
\item Bound estimation and state-construction error separately; certificates inherit both.
\end{enumerate}

\paragraph{Limitations.}
The guarantees assume fixed supported rows, rewards paid on absorption, policies that are proper without discounting
and a sink that is a worst absorbing state. NHB cannot be checked from the released rows; the NHB-free test and the
breakdown value exist for this reason. Ground truth exists only in our tabular simulators, so the validation reaches
\icu{} by analogy. On the Gumbel simulator most decision states admit a single supported action, which favours
termination; softer physicians reduce this share without producing errors, but the simulators remain simpler than
real care. Dataset draws are replicates, not independent hospitals. Apart from H1--H3, which were fixed in code but not
externally registered, our analyses are exploratory. All reported numbers come from one uninterrupted run of our code
from an empty folder (\cref{app:repro}).

\paragraph{Conclusion.}
Filling in unsupported actions is a modelling choice, and the two standard rules mark its limits. Termination bounds
every completion from below, and the mean rule bounds those without hidden benefit from above. Reporting this interval, and certifying
only comparisons that survive it, removes the completion rule as a hidden source of conclusions. It does not remove
estimation or aggregation error, which is the next problem for benchmarks built from records.

\section*{Broader impact statement}
We use no patient-level data; \icu{} is derived from de-identified MIMIC-III records. The terminate rule is a stress
test. An unsupported action is one that was rarely given, which does not make it unsafe. Benchmark returns, including
ours, are not clinical evidence, and no policy in this paper is fit for clinical use.

\ifpreprint
\section*{Acknowledgements}
This work began as a project in the Reinforcement Learning course (CSE753) at BRAC University, taught by
Dr.\ Swakkhar Shatabda.
\fi

\section*{Code availability}
Code, exported results and scripts that rebuild every table, figure and number are available at
\url{https://github.com/AshrafulKabir7/certified-comparisons-offline-rl}.

\section*{LLM usage}
LLM assistants were used for drafting and editing text and for writing analysis code. The authors verified all content
and take full responsibility for it.

\bibliographystyle{plainnat}
\bibliography{refs}

\appendix
\crefalias{section}{appendix}

\section{Proofs}\label{app:proofs}

Throughout, rows $P_0$ and $P_1$ are \emph{compatible} if they share the state space, rewards, initial distribution,
support sets and absorbing rows and agree on every supported live pair; all completions of one benchmark are pairwise
compatible. For a policy $\pi$ write $P^\pi(s'\mid s)=\sum_a\pi(a\mid s)P(s'\mid s,a)$ and
$b_\pi(s)=\sum_{a\notin\M(s)}\pi(a\mid s)$.

\paragraph{Standard facts.}
(F1) If $\pi$ is proper under $P$, then $F=I-P^\pi[\Sset,\Sset]$ is invertible, $F^{-1}=\sum_{k\ge0}(P^\pi[\Sset,\Sset])^k$
is entrywise nonnegative and $\mu_P^\pi{}^\top=d_0^\top F^{-1}$.
(F2) For proper $\pi$, $v_P^\pi(s)$ is the expected reward of the absorbing state eventually reached from $s$, hence a
convex combination of $\{R(t):t\in\Tset\}$. Therefore $U_P^\pi$ takes values in $[\min_\Tset R,\Rmax]\subseteq[\vmin,\vmax]$.
(F3) Histories up to the first unsupported decision use only supported rows, so $p_\pi$ is the same under every
completion. Under $\Pterm$ the first unsupported decision ends the episode, so
$p_\pi=\sum_{s}\mu^\pi_{\Pterm}(s)\,b_\pi(s)$.
(F4) If $\pi$ is proper under a completion $P$, it is proper under $\Pterm$ and under every completion whose unsupported
rows are mixtures of $P$'s unsupported rows and absorbing states. Couple the two chains on the same randomness: they
coincide until an unsupported decision, at which point the second chain either continues as the first or is absorbed.

\begin{lemma}[Simulation identity]\label{lem:sim}
Let $P_0,P_1$ be compatible and let $\pi$ be proper under both. Then \eqref{eq:sim} holds.
\end{lemma}

\begin{proof}
Write $v_i=v_{P_i}^\pi$ and $U_i=U_{P_i}^\pi$. Because rewards are paid only on absorption and $v_i=0$ on $\Tset$, every
live $s$ satisfies $v_i(s)=\sum_{s'}P_i^\pi(s'\mid s)U_i(s')$. Subtracting the two equations,
\[
v_0(s)-v_1(s)=\underbrace{\textstyle\sum_{s'}\big(P_0^\pi-P_1^\pi\big)(s'\mid s)\,U_0(s')}_{=:\Delta(s)}
+\textstyle\sum_{s'}P_1^\pi(s'\mid s)\big(U_0-U_1\big)(s').
\]
$U_0-U_1$ vanishes on $\Tset$ and equals $v_0-v_1$ on $\Sset$, so $(I-P_1^\pi[\Sset,\Sset])(v_0-v_1)=\Delta$. By (F1),
$J_{P_0}(\pi)-J_{P_1}(\pi)=d_0^\top(v_0-v_1)=\mu_{P_1}^\pi{}^\top\Delta$. The rows agree on supported pairs, so only
unsupported actions contribute to $\Delta(s)$, which gives \eqref{eq:sim}.
\end{proof}

\begin{proof}[Proof of \cref{thm:bracket}]
(i) Apply \cref{lem:sim} with $P_0=P^\star$ and $P_1=\Pterm$, which is allowed by (F4). For unsupported $a$,
$\Pterm(\cdot\mid s,a)$ is the point mass on $s^\dagger$, so the contribution of $(s,a)$ is
$\mu^\pi_{\Pterm}(s)\pi(a\mid s)\big(P^\star(\cdot\mid s,a)^\top U_{P^\star}^\pi-\rterm\big)$. By (F2),
$P^\star(\cdot\mid s,a)^\top U_{P^\star}^\pi\ge\vmin=\rterm$, so every contribution is nonnegative.

(ii) Apply \cref{lem:sim} with $P_0=P_W$ and $P_1=P^\star$:
$\JW(\pi)-J_{P^\star}(\pi)=\sum_s\mu^\pi_{P^\star}(s)\sum_{a\notin\M(s)}\pi(a\mid s)\big(P_W-P^\star\big)(\cdot\mid s,a)^\top
U_{P_W}^\pi$, and each term is nonnegative by NHB.

(iii) Apply \cref{lem:sim} with $P_0=P_W$ and $P_1=\Pterm$:
\[
\JW(\pi)-\Jterm(\pi)=\sum_s\mu^\pi_{\Pterm}(s)\sum_{a\notin\M(s)}\pi(a\mid s)\big(P_W(\cdot\mid s,a)^\top U^\pi_{P_W}-\rterm\big).
\]
By (F2) each bracket lies in $[0,\vmax-\vmin]$, and by (F3) the weights sum to $p_\pi$. If $p_\pi=0$, all weights
$\mu^\pi_{\Pterm}(s)\pi(a\mid s)$ on unsupported pairs vanish, so \cref{lem:sim} with $P_1=\Pterm$ gives
$J_{P}(\pi)=\Jterm(\pi)$ for every completion $P$ under which $\pi$ is proper.

(iv) For $\delta\in[0,1]$ let $P_\delta$ have unsupported rows $(1-\delta)P_W(\cdot\mid s,a)+\delta\,\mathbb{1}_{s^\dagger}$.
By (F4), $\pi$ is proper under every $P_\delta$. The entries of $P_\delta^\pi$ are affine in $\delta$ and
$I-P_\delta^\pi[\Sset,\Sset]$ is invertible on $[0,1]$, so $\delta\mapsto J_{P_\delta}(\pi)$ is continuous, with
$J_{P_0}(\pi)=\JW(\pi)$ and $J_{P_1}(\pi)=\Jterm(\pi)$. By the intermediate value theorem every value in
$[\Jterm(\pi),\JW(\pi)]$ is attained. Each $P_\delta$, and more generally every intermediate-harm completion, has NHB
for $\pi$ because
$P_\delta(\cdot\mid s,a)^\top U^\pi_{P_W}=(1-\delta)P_W(\cdot\mid s,a)^\top U^\pi_{P_W}+\delta\rterm
\le P_W(\cdot\mid s,a)^\top U^\pi_{P_W}$, using $P_W(\cdot\mid s,a)^\top U^\pi_{P_W}\ge\vmin=\rterm$. Together with (i) and
(ii), the set of values of $\pi$ under NHB, and under intermediate harm, is exactly $[\Jterm(\pi),\JW(\pi)]$.

Without NHB, the computation in (i) and $U_{P}^\pi\le\Rmax$ (F2) give, for every completion $P$ under which $\pi$ is
proper,
\[
J_P(\pi)-\Jterm(\pi)\le\sum_s\mu^\pi_{\Pterm}(s)\,b_\pi(s)\,(\Rmax-\rterm)=p_\pi(\Rmax-\vmin).
\] Let $t^+\in\arg\max_\Tset R$ and, for $\lambda\in[0,1]$, let $P^\lambda$ send each unsupported action to
$t^+$ with probability $\lambda$ and to $s^\dagger$ otherwise. Since both targets are absorbing,
\cref{lem:sim} with $P_1=\Pterm$ gives $J_{P^\lambda}(\pi)-\Jterm(\pi)=\lambda\,p_\pi(\Rmax-\vmin)$, which fills the stated
range.
\end{proof}

\begin{proof}[Proof of \cref{cor:cert}]
If $P^\star$ has NHB for $\pi_2$, \cref{thm:bracket}(i) for $\pi_1$ and (ii) for $\pi_2$ give
$J_{P^\star}(\pi_1)\ge\Jterm(\pi_1)>\JW(\pi_2)\ge J_{P^\star}(\pi_2)$; nothing is assumed about $\pi_1$. For the NHB-free
test, \cref{thm:bracket}(iv) gives $J_{P^\star}(\pi_2)\le\Jterm(\pi_2)+p_{\pi_2}(\Rmax-\vmin)$ for every completion. In both
cases the single rules agree: $\JW(\pi_1)\ge\Jterm(\pi_1)>\JW(\pi_2)\ge\Jterm(\pi_2)$ by (iii), where for the NHB-free test
$\JW(\pi_2)\le\Jterm(\pi_2)+p_{\pi_2}(\Rmax-\vmin)$ holds because $P_W$ is itself a completion.
\end{proof}

\paragraph{The sensitivity family.}
$E(\pi)\ge0$ because $P_W$ is a completion, so $\JW(\pi)\le\Jterm(\pi)+p_\pi(\Rmax-\vmin)$ by \cref{thm:bracket}(iv). If
\eqref{eq:kappa} holds and a completion satisfies $J_{P^\star}(\pi_2)\le\JW(\pi_2)+\kappa E(\pi_2)$, then
$J_{P^\star}(\pi_1)\ge\Jterm(\pi_1)>\JW(\pi_2)+\kappa E(\pi_2)\ge J_{P^\star}(\pi_2)$. At $\kappa=1$ the right-hand side of
\eqref{eq:kappa} equals $\Jterm(\pi_2)+p_{\pi_2}(\Rmax-\vmin)$, the NHB-free test, and every completion satisfies the
condition. The breakdown value of a claim is $\kappa^*=\sup\{\kappa\in[0,1]:\Jterm(\pi_1)>\JW(\pi_2)+\kappa E(\pi_2)\}$.

\begin{proposition}[Completed benchmarks make support constraints free]\label{prop:flat}
Assume, as in \cref{sec:setting}, that the optima below are attained by proper stationary policies.
(i) Under $P_W$, every policy $\pi$ and its $W$-projection
$\pi_W(a\mid s)=\pi(a\mid s)\mathbb{1}[a\in\M(s)]+\sum_{a'\notin\M(s)}\pi(a'\mid s)W(a\mid s,a')$ induce the same chain, so
they have the same properness and value. (ii) Under $\Pterm$ with $\rterm=\vmin$, no proper policy is worth more than the
best admissible one. In both cases the violation-constrained frontier $\eta\mapsto\max\{J(\pi):\mathbb{E}[N_{\mathrm{viol}}]\le\eta\}$
is constant and equal to the admissible optimum.
\end{proposition}

\begin{proof}
(i) For every live $s$, $P_W^\pi(\cdot\mid s)=\sum_{a\in\M(s)}\pi(a\mid s)P(\cdot\mid s,a)+\sum_{a'\notin\M(s)}\pi(a'\mid s)
\sum_{a\in\M(s)}W(a\mid s,a')P(\cdot\mid s,a)=P_W^{\pi_W}(\cdot\mid s)$. Since $\pi_W$ is admissible,
$J(\pi)=J(\pi_W)$ is at most the admissible optimum. (ii) Let $v_\M$ be the optimal value among admissible policies and
$U_\M=R+v_\M$, so that $v_\M(s)=\max_{a\in\M(s)}P(\cdot\mid s,a)^\top U_\M$. An unsupported action leads to $s^\dagger$ and is
worth $\rterm=\vmin\le P(\cdot\mid s,a')^\top U_\M$ for every supported $a'$ (F2). Hence $v_\M\ge P^\pi U_\M$ on $\Sset$ for
every policy $\pi$. Subtracting $v_\pi=P^\pi U_\pi$ and using that $U_\M-U_\pi$ vanishes on $\Tset$ gives
$(I-P^\pi[\Sset,\Sset])(v_\M-v_\pi)\ge0$, so $v_\M\ge v_\pi$ by (F1) for every proper $\pi$. The frontier statements follow
because admissible policies have $\mathbb{E}[N_{\mathrm{viol}}]=0$.
\end{proof}

\section{Environments and benchmark construction}\label{app:env}

\paragraph{Properness.}
Absorption hazards (the per-step timeout on the Gumbel simulator) and, on Garnet, censoring of trap states make every
policy proper in our simulators. No \icu{} evaluation under either rule is improper.

\paragraph{The Gumbel-max sepsis simulator as an exact MDP.}
We enumerate the simulator of \citet{oberst2019gumbel} exactly from its published code: 606 live states and 8 actions
(combinations of antibiotics, ventilation and vasopressors). Death pays $-1$, discharge pays $+1$, and a per-step
timeout of probability 0.05 makes every policy proper; it ranks policies like discounting by 0.95. Exact evaluation
agrees with Monte Carlo simulation of the original code.

\paragraph{Cells.}
\cref{tab:envs} lists the main cells. Garnet benchmarks are built from 5,000 hospital episodes with $\tau=5$. The
\emph{oracle} construction keeps the true state space, initial distribution and supported rows, so its only error is
the completion. A benchmark policy is lifted to the true state space by copying it on the benchmark's decision states
and playing the most frequent hospital action elsewhere.

\begin{table}[htbp]
\centering
\caption{Environment cells, draw 0. est.: ICU-style construction; oracle: true states and supported rows. hosp.:
hospital episodes; $\overline{|M(s)|}$: mean number of supported actions per live state; $J_\beta^{\mathrm{true}}$:
true value of the physician; $J^*_{\mathrm{true}}$, $J^*_{\mathrm{bench}}$: optimal value in truth and on the
benchmark; $V_{\mathrm{unsupp}}$: gain of the true optimum over the true optimum restricted to the mask. \icu{} has no
ground truth: its $J_\beta^{\mathrm{true}}$ entry is the clinician's value under the terminate rule, the deployment
target, and its $J^*_{\mathrm{bench}}$ entry is the benchmark optimum.}
\label{tab:envs}
\resizebox{\linewidth}{!}{\begin{tabular}{llrrrrrrrr}
\toprule
Cell & constr. & hosp. & $\tau$ & live $|\mathcal S|$ & $\overline{|M(s)|}$ & $J_{\beta}^{\mathrm{true}}$ & $J^{*}_{\mathrm{true}}$ & $J^{*}_{\mathrm{bench}}$ & $V_{\mathrm{unsupp}}$ \\
\midrule
ICU-Sepsis v2 & released & -- & -- & 713 & 3.14 & 0.416 & -- & 0.875 & -- \\
Gumbel: $n{=}2$k, $\tau{=}5$ & est. & 2000 & 5 & 184 & 1.27 & 0.350 & 0.437 & 0.478 & 0.000 \\
Gumbel: $n{=}2$k, $\tau{=}20$ & est. & 2000 & 20 & 87 & 1.00 & 0.318 & 0.437 & 0.474 & 0.000 \\
Gumbel: $n{=}10$k, $\tau{=}5$ & est. & 10000 & 5 & 290 & 1.68 & 0.356 & 0.437 & 0.475 & 0.000 \\
Gumbel: $n{=}10$k, $\tau{=}20$ & est. & 10000 & 20 & 192 & 1.25 & 0.350 & 0.437 & 0.443 & 0.000 \\
Gumbel: $n{=}10$k, $\tau{=}20$, $\epsilon{=}0.2$ & est. & 10000 & 20 & 242 & 1.64 & 0.150 & 0.437 & 0.467 & 0.000 \\
Gumbel: $n{=}10$k, $\tau{=}20$, withhold-vaso & est. & 10000 & 20 & 201 & 1.46 & 0.158 & 0.437 & 0.345 & 0.117 \\
Gumbel: oracle & oracle & 10000 & 20 & 606 & 2.99 & 0.358 & 0.437 & 0.437 & 0.000 \\
Garnet: $h{=}{-}0.5$, $\epsilon{=}0.05$ & est. & 5000 & 5 & 150 & 2.74 & 0.870 & 0.986 & 0.997 & 0.007 \\
Garnet: $h{=}{-}0.5$, $\epsilon{=}0.3$ & est. & 5000 & 5 & 150 & 5.14 & 0.939 & 0.986 & 1.000 & 0.001 \\
Garnet: $h{=}0.0$, $\epsilon{=}0.05$ & est. & 5000 & 5 & 150 & 3.19 & 0.841 & 0.915 & 0.989 & 0.000 \\
Garnet: $h{=}0.0$, $\epsilon{=}0.3$ & est. & 5000 & 5 & 149 & 7.17 & 0.843 & 0.915 & 1.000 & 0.000 \\
Garnet: $h{=}0.5$, $\epsilon{=}0.05$ & est. & 5000 & 5 & 150 & 2.68 & 0.690 & 0.915 & 0.939 & 0.000 \\
Garnet: $h{=}0.5$, $\epsilon{=}0.3$ & est. & 5000 & 5 & 150 & 5.03 & 0.338 & 0.915 & 0.954 & 0.000 \\
Garnet: $h{=}1.0$, $\epsilon{=}0.05$ & est. & 5000 & 5 & 150 & 2.55 & 0.576 & 0.915 & 0.932 & 0.000 \\
Garnet: $h{=}1.0$, $\epsilon{=}0.3$ & est. & 5000 & 5 & 137 & 3.51 & 0.183 & 0.915 & 0.971 & 0.001 \\
\bottomrule
\end{tabular}
}
\end{table}

\paragraph{Additional cells.}
Four Gumbel cells change the physician or the state space (\cref{tab:physicians,tab:addcells}). A confounded physician
gives diabetic patients the non-diabetic optimum. A second physician withholds antibiotics, so the optimal action lies
in only 17\% of draw-0 masks. Two cells estimate the benchmark on aggregated states, as \icu{} does on clusters: the
observation without diabetes, a confounder the physician sees, or without diabetes and glucose. Two Garnet cells draw
every non-preferred row independently, so whether an unsupported action helps varies by pair. Further sensitivity
cells soften the Gumbel physician ($\epsilon\in\{0.3,0.5\}$) or vary the timeout hazard, the dead-state rule and the
state abstraction; they are reported in \cref{tab:sens} (\cref{app:results}).

\begin{table}[htbp]
\centering
\caption{The four Gumbel physicians on the true system: value, value of the deterministic core ($\epsilon=0$), share of
live states (unweighted, weighted by the physician's occupancy, and diabetic states only) where the main action
differs from the true optimum, and share of live states whose main action includes each treatment.}
\label{tab:physicians}
\resizebox{\linewidth}{!}{\begin{tabular}{lrrrrrrrr}
\toprule
 & & & \multicolumn{3}{c}{disagrees with optimum (\%)} & \multicolumn{3}{c}{treatment share (\%)} \\
\cmidrule(lr){4-6}\cmidrule(lr){7-9}
Physician ($\epsilon{=}0.05$) & value & core & live & occup. & diabetic & abx & vent & vaso \\
\midrule
$\epsilon$-optimal (main cells) & 0.3581 & 0.4365 & 0.00 & 0.00 & 0.00 & 87.62 & 96.53 & 26.40 \\
withhold-vasopressor (control) & 0.1595 & 0.2336 & 26.57 & 33.14 & 21.12 & 90.10 & 96.70 & 0.00 \\
confounded (diabetes-blind) & 0.3301 & 0.3998 & 6.77 & 5.04 & 13.53 & 86.47 & 97.69 & 31.68 \\
withhold-antibiotics & $-$0.2346 & $-$0.2014 & 87.95 & 72.77 & 89.44 & 0.00 & 90.76 & 29.70 \\
\bottomrule
\end{tabular}
}
\end{table}

\begin{table}[htbp]
\centering
\caption{Additional cells. Benchmark of draw 0: live states, mean mask size, share of decision states with one supported
action, share whose mask contains the true optimal action, and $V_{\mathrm{unsupp}}$. True values (means over 30
draws) of the benchmark's clinician policy $\hat\pi_\beta$, projected IQL and the mask-restricted optimum
$J^*_{\M}$.}
\label{tab:addcells}
\resizebox{\linewidth}{!}{\begin{tabular}{lrrrrrrrr}
\toprule
 & \multicolumn{5}{c}{benchmark of draw 0} & \multicolumn{3}{c}{true value (30 draws)} \\
\cmidrule(lr){2-6}\cmidrule(lr){7-9}
Cell & live & mean $|\M|$ & $|\M|{=}1$ & opt.\ $\in\M$ & $V_{\mathrm{unsupp}}$ & $\pibhat$ & proj.\ IQL & $J^*_{\M}$ \\
\midrule
Gumbel, confounded physician & 189 & 1.30 & 94.7 & 95.8 & 0.0345 & 0.3239 & 0.3963 & 0.4112 \\
Gumbel, withhold-abx physician & 180 & 1.12 & 98.3 & 16.7 & 0.4993 & $-$0.2579 & $-$0.2099 & $-$0.0665 \\
Gumbel, aggregated, diabetes hidden & 133 & 1.37 & 90.2 & 94.6 & 0.0003 & 0.3208 & 0.4008 & 0.4359 \\
Gumbel, aggregated, diabetes \& glucose hidden & 53 & 1.94 & 71.7 & 94.9 & 0.0010 & 0.0908 & $-$0.0057 & 0.4361 \\
Garnet, random rows, $\epsilon{=}0.05$ & 150 & 3.00 & 22.7 & -- & 0.0209 & 0.8298 & 0.8642 & 0.9202 \\
Garnet, random rows, $\epsilon{=}0.3$ & 150 & 7.59 & 0 & -- & 0.0014 & 0.7353 & 0.8603 & 0.9429 \\
\bottomrule
\end{tabular}
}
\end{table}

\section{Methods and policies}\label{app:methods}

\paragraph{The tabular suite.}
All 20 methods (\cref{sec:validation}) run exactly on the estimated benchmark and use 80\% of each user dataset. Each
yields a native policy and its projection with uniform $W$; eight also yield a masked extraction, which restricts the
greedy step to supported actions. Together with the behaviour policy this gives 50 default policies per draw. Each
method has one tuned hyper-parameter whose default was declared before the run; grids, defaults and convergence
handling are documented in the code release.

\paragraph{The 19 fixed \icu{} policies.}
These policies were trained on one frozen dataset of 10,000 simulated clinician episodes with 10 training seeds, fixed
before the certificates were computed, and evaluated exactly on the released rows (\cref{tab:icumain}). They include
tabular BCQ, IQL and CQL with masked, unmasked and projected extraction; advantage-weighted categorical imitation; deep
CQL, critic ensembles and CQL variants with hard or soft action masks; critic-guided diffusion policies; the
clinician; a uniformly random policy; and the optimum, which is computed from the model and is not an offline
competitor.

The neural variants in \cref{tab:icumain} differ only in how a CQL-style critic treats unsupported actions. Deep CQL
regularises with a log-sum-exp over all actions and exports an unmasked or masked greedy policy. The hard
action-masked variants back up and export only supported actions; they differ in the regulariser. Design A takes the
log-sum-exp over all actions on every logged row. Design B takes it over supported actions only, on rows whose logged
action is supported. The legacy design takes it over supported actions on every row, including rows whose logged
action is unsupported, and is kept as a failure-mode control. The soft variant keeps all actions and adds a dual
penalty on the softmax mass the critic puts on unsupported actions. The masked critic ensemble is the masked greedy
policy of an ensemble of design-B critics. The two diffusion policies are categorical diffusion models whose sampling
is restricted to supported actions and guided by a critic: the design-B critic (critic B) or the legacy critic.
Advantage-weighted categorical imitation fits a classifier to the logged actions with weights from tabular IQL
advantages. Projected tabular IQL is the unconstrained tabular IQL policy made admissible by the uniform projection
of \cref{prop:flat}.

\begin{table}[htbp]
\centering
\caption{The 19 fixed policies on \icu{} v2: exact returns under the mean and terminate rules, mean $\pm$ s.d.\ over
training seeds. viol.: share of expected decisions that are unsupported. Top block: policies with no unsupported mass,
for which the two rules agree exactly. Middle: unmasked tabular policies. Bottom: references.}
\label{tab:icumain}
\small
\begin{tabular}{lrrrr}
\toprule
Policy & seeds & $J_{\mathrm{mean}}$ & $J_{\mathrm{term}}$ & viol.\ (\%) \\
\midrule
Tabular BCQ, masked greedy & 10 & 0.8034\,$\pm$\,0.0016 & 0.8034\,$\pm$\,0.0016 & 0.00 \\
Tabular IQL, masked greedy & 10 & 0.7909\,$\pm$\,0.0019 & 0.7909\,$\pm$\,0.0019 & 0.00 \\
Tabular IQL, projected & 10 & 0.7906\,$\pm$\,0.0020 & 0.7906\,$\pm$\,0.0020 & 0.00 \\
AW categorical imitation, masked & 10 & 0.7899\,$\pm$\,0.0031 & 0.7899\,$\pm$\,0.0031 & 0.00 \\
Tabular CQL, masked greedy & 10 & 0.7855\,$\pm$\,0.0019 & 0.7855\,$\pm$\,0.0019 & 0.00 \\
Diffusion, legacy critic & 10 & 0.7850\,$\pm$\,0.0031 & 0.7850\,$\pm$\,0.0031 & 0.00 \\
Diffusion, critic B & 10 & 0.7843\,$\pm$\,0.0011 & 0.7843\,$\pm$\,0.0011 & 0.00 \\
Masked critic ensemble (no diffusion) & 10 & 0.7843\,$\pm$\,0.0015 & 0.7843\,$\pm$\,0.0015 & 0.00 \\
Action-masked CQL (hard), design B & 10 & 0.7821\,$\pm$\,0.0013 & 0.7821\,$\pm$\,0.0013 & 0.00 \\
Action-masked CQL (hard), legacy & 10 & 0.7821\,$\pm$\,0.0043 & 0.7821\,$\pm$\,0.0043 & 0.00 \\
Action-masked CQL (hard), design A & 10 & 0.7783\,$\pm$\,0.0025 & 0.7783\,$\pm$\,0.0025 & 0.00 \\
Deep CQL, masked greedy & 10 & 0.7789\,$\pm$\,0.0031 & 0.7789\,$\pm$\,0.0031 & 0.00 \\
Action-masked CQL (soft) & 10 & 0.7766\,$\pm$\,0.0020 & 0.7766\,$\pm$\,0.0020 & 0.00 \\
\midrule
Tabular IQL, unmasked & 10 & 0.7906\,$\pm$\,0.0020 & 0.7801\,$\pm$\,0.0033 & 0.30 \\
Tabular CQL, unmasked & 10 & 0.7848\,$\pm$\,0.0020 & 0.7623\,$\pm$\,0.0032 & 0.71 \\
Tabular BCQ, unmasked & 10 & 0.8015\,$\pm$\,0.0018 & 0.7050\,$\pm$\,0.0096 & 3.42 \\
\midrule
Clinician (behaviour) & 1 & 0.7818 & 0.4158 & 12.10 \\
Uniform random & 1 & 0.7801 & 0.0084 & 85.87 \\
Optimal (= constrained optimal) & 1 & 0.8751 & 0.8751 & 0.00 \\
\bottomrule
\end{tabular}

\end{table}

\section{Additional results}\label{app:results}

\paragraph{Per-cell results.}
\cref{tab:certified} reports the bounds and certificates for every cell, and \cref{tab:pairtypes} splits the
certified claims by how many of the two policies can leave the support.

\begin{table}[htbp]
\centering
\caption{Bounds and certified comparisons per cell (30 draws per cell). lower: lower-bound violations; upper, hold: \%
of policy-draws with $p_\pi>0$ (all 110 policies per draw) within the upper bound or the interval; share: \% of pairs certified; err.: \% of
certified claims the truth reverses or ties; point err.: the same for strict single-rule claims (mean/terminate).}
\label{tab:certified}
\resizebox{\linewidth}{!}{\setlength{\tabcolsep}{3pt}%
\begin{tabular}{lrrrrcrrrc}
\toprule
 & \multicolumn{5}{c}{semantics only (true supported rows)} & \multicolumn{4}{c}{benchmark (estimated rows)} \\
\cmidrule(lr){2-6}\cmidrule(lr){7-10}
 & \multicolumn{2}{c}{bounds} & \multicolumn{2}{c}{certified} & point err. & bounds & \multicolumn{2}{c}{certified} & point err. \\
\cmidrule(lr){2-3}\cmidrule(lr){4-5}\cmidrule(lr){6-6}\cmidrule(lr){7-7}\cmidrule(lr){8-9}\cmidrule(lr){10-10}
Cell & lower & upper & share & err. & mean/term & hold & share & err. & mean/term \\
\midrule
\multicolumn{10}{l}{\textit{Gumbel-max sepsis} (hospital episodes, threshold $\tau$)} \\
2k, $\tau{=}5$ & 0 & 99.4 & 73.5 & 0 & 6.4/2.9 & 85.0 & 72.5 & 23.2 & 30.1/21.4 \\
2k, $\tau{=}20$ & 0 & 100.0 & 42.8 & 0 & 10.7/0.4 & 85.3 & 43.1 & 6.2 & 15.8/4.8 \\
10k, $\tau{=}5$ & 0 & 99.7 & 78.4 & 0 & 2.8/3.5 & 55.5 & 77.4 & 7.9 & 10.8/10.8 \\
10k, $\tau{=}20$ & 0 & 99.9 & 71.9 & 0 & 7.4/2.4 & 81.7 & 72.0 & 7.5 & 14.7/8.4 \\
10k, $\tau{=}20$, $\epsilon{=}0.2$ & 0 & 99.9 & 75.9 & 0 & 6.6/2.4 & 77.5 & 76.1 & 4.6 & 10.5/6.4 \\
10k, $\tau{=}20$, withhold-vaso & 0 & 94.5 & 71.8 & 0 & 10.9/3.6 & 83.0 & 70.8 & 12.7 & 23.3/13.3 \\
oracle, 10k, $\tau{=}20$ & 0 & 97.7 & 75.5 & 0 & 8.5/2.5 & 97.7 & 75.5 & 0 & 8.5/2.5 \\
pooled (6 estimated cells) & 0 & 99.0 & 69.0 & 0 & 7.2/2.6 & 76.4 & 68.6 & 10.6 & 17.6/11.1 \\
\midrule
\multicolumn{10}{l}{\textit{Garnet}} \\
$h{=}{-}0.5$, $\epsilon{=}0.05$ & 0 & 1.9 & 79.6 & 4.8 & 4.0/20.9 & 36.9 & 82.2 & 12.9 & 11.3/25.0 \\
$h{=}{-}0.5$, $\epsilon{=}0.3$ & 0 & 1.5 & 78.3 & 3.2 & 2.7/21.0 & 53.7 & 79.6 & 8.0 & 7.7/23.1 \\
$h{=}0.0$, $\epsilon{=}0.05$ & 0 & 85.8 & 81.8 & $<$0.1 & $<$0.1/14.8 & 75.2 & 83.2 & 14.5 & 14.2/24.1 \\
$h{=}0.0$, $\epsilon{=}0.3$ & 0 & 77.1 & 80.0 & $<$0.1 & 0.2/15.9 & 55.0 & 84.4 & 17.7 & 17.5/25.2 \\
$h{=}0.5$, $\epsilon{=}0.05$ & 0 & 96.3 & 90.4 & 0 & 3.5/0.8 & 75.0 & 90.2 & 4.6 & 7.9/5.5 \\
$h{=}0.5$, $\epsilon{=}0.3$ & 0 & 99.8 & 91.3 & 0 & 3.1/2.0 & 65.2 & 91.1 & 4.6 & 7.6/6.4 \\
$h{=}1.0$, $\epsilon{=}0.05$ & 0 & 97.1 & 90.2 & $<$0.1 & 4.2/0.3 & 22.9 & 90.3 & 4.0 & 8.0/3.9 \\
$h{=}1.0$, $\epsilon{=}0.3$ & 0 & 99.0 & 88.4 & $<$0.1 & 6.5/1.0 & 18.0 & 89.1 & 8.5 & 14.2/8.6 \\
pooled (8 cells) & 0 & 69.0 & 85.0 & 1.0 & 3.0/9.6 & 49.3 & 86.3 & 9.2 & 11.0/15.2 \\
\midrule
\multicolumn{10}{l}{\textit{Additional cells} (Gumbel: 10k, $\tau{=}20$, $\epsilon{=}0.05$; Garnet: random unsupported rows)} \\
confounded phys. & 0 & 96.4 & 72.6 & 0 & 6.9/2.6 & 80.3 & 72.5 & 6.6 & 13.5/8.0 \\
withhold-abx phys. & 0 & 97.5 & 68.8 & 0 & 8.5/6.9 & 73.8 & 66.4 & 7.9 & 17.6/13.0 \\
aggregated, diabetes hidden & -- & -- & -- & -- & -- & 76.3 & 72.5 & 43.7 & 51.6/36.6 \\
aggregated, diabetes \& glucose hidden & -- & -- & -- & -- & -- & 23.3 & 85.5 & 9.8 & 12.8/12.2 \\
random rows, $\epsilon{=}0.05$ & 0 & 97.7 & 83.2 & $<$0.1 & 4.5/7.8 & 86.2 & 82.8 & 14.3 & 18.3/19.0 \\
random rows, $\epsilon{=}0.3$ & 0 & 84.8 & 92.3 & $<$0.1 & 0.2/3.1 & 47.6 & 93.6 & 11.9 & 12.5/13.0 \\
\midrule
ICU-Sepsis (no truth) & -- & -- & -- & -- & -- & -- & 72.9 & -- & -- \\
\bottomrule
\end{tabular}
}
\end{table}

\begin{table}[htbp]
\centering
\caption{Certified comparisons by pair type: neither, one or both policies make an unsupported decision with positive
probability, and after merging policies with identical values (distinct). Share of pairs certified and share of claims
the truth reverses or ties (\%); point err.: single-rule claims.}
\label{tab:pairtypes}
\resizebox{\linewidth}{!}{\setlength{\tabcolsep}{3pt}%
\begin{tabular}{llrrrrrrr}
\toprule
 & & & \multicolumn{2}{c}{NHB certificate} & \multicolumn{2}{c}{NHB-free certificate} & \multicolumn{2}{c}{point err.} \\
\cmidrule(lr){4-5}\cmidrule(lr){6-7}\cmidrule(lr){8-9}
Cells & pair type & pairs & share & err. & share & err. & mean & term \\
\midrule
\multicolumn{9}{l}{\textit{true supported rows}} \\
Gumbel (6 estimated cells) & neither with $p_\pi{>}0$ & 138{,}488 & 73.2 & 0 & 73.2 & 0 & 0 & 0 \\
 & one with $p_\pi{>}0$ & 73{,}080 & 66.6 & 0 & 11.9 & 0 & 17.0 & 5.4 \\
 & both with $p_\pi{>}0$ & 8{,}932 & 23.6 & 0 & 2.3 & 0 & 21.6 & 10.7 \\
 & distinct policies & 49{,}276 & 70.8 & 0 & 46.9 & 0 & 14.9 & 5.8 \\
\addlinespace[2pt]
Garnet $h{=}{-}0.5$ & neither with $p_\pi{>}0$ & 41{,}384 & 97.4 & 0 & 97.4 & 0 & 0 & 0 \\
 & one with $p_\pi{>}0$ & 27{,}629 & 59.4 & 14.0 & 47.7 & 0 & 8.6 & 48.9 \\
 & both with $p_\pi{>}0$ & 4{,}487 & 29.3 & 3.3 & 24.3 & 0 & 2.1 & 37.2 \\
 & distinct policies & 37{,}098 & 77.1 & 4.8 & 71.3 & 0 & 3.9 & 23.5 \\
\addlinespace[2pt]
Garnet $h{=}0$ & neither with $p_\pi{>}0$ & 46{,}535 & 96.7 & 0 & 96.7 & 0 & 0 & 0 \\
 & one with $p_\pi{>}0$ & 24{,}040 & 58.2 & 0.3 & 24.7 & 0 & 0.3 & 41.0 \\
 & both with $p_\pi{>}0$ & 2{,}925 & 16.7 & 0 & 7.9 & 0 & 0 & 40.2 \\
 & distinct policies & 29{,}743 & 75.6 & 0.1 & 61.8 & 0 & 0.2 & 20.7 \\
\addlinespace[2pt]
Garnet $h{\ge}0.5$ & neither with $p_\pi{>}0$ & 85{,}153 & 97.2 & 0 & 97.2 & 0 & 0 & 0 \\
 & one with $p_\pi{>}0$ & 53{,}526 & 83.6 & $<$0.1 & 63.2 & 0 & 10.9 & 2.5 \\
 & both with $p_\pi{>}0$ & 8{,}321 & 58.9 & 0 & 37.6 & 0 & 6.7 & 2.1 \\
 & distinct policies & 70{,}871 & 88.5 & $<$0.1 & 78.4 & 0 & 5.9 & 1.3 \\
\midrule
\multicolumn{9}{l}{\textit{estimated rows (benchmark)}} \\
Gumbel (6 estimated cells) & neither with $p_\pi{>}0$ & 138{,}986 & 73.2 & 14.4 & 73.2 & 14.4 & 14.4 & 14.4 \\
 & one with $p_\pi{>}0$ & 72{,}721 & 65.5 & 2.7 & 11.2 & 5.3 & 21.1 & 6.6 \\
 & both with $p_\pi{>}0$ & 8{,}793 & 21.9 & 1.7 & 1.6 & 0 & 30.9 & 10.5 \\
 & distinct policies & 49{,}276 & 70.6 & 9.6 & 46.7 & 12.8 & 23.7 & 12.0 \\
\addlinespace[2pt]
Garnet $h{=}{-}0.5$ & neither with $p_\pi{>}0$ & 41{,}865 & 97.4 & 6.9 & 97.4 & 6.9 & 6.9 & 6.9 \\
 & one with $p_\pi{>}0$ & 27{,}304 & 63.1 & 19.3 & 51.4 & 4.4 & 13.8 & 47.8 \\
 & both with $p_\pi{>}0$ & 4{,}331 & 33.0 & 6.0 & 28.5 & 1.7 & 7.1 & 35.1 \\
 & distinct policies & 37{,}098 & 79.2 & 10.6 & 73.9 & 5.8 & 9.6 & 25.6 \\
\addlinespace[2pt]
Garnet $h{=}0$ & neither with $p_\pi{>}0$ & 46{,}918 & 96.7 & 16.8 & 96.7 & 16.8 & 16.8 & 16.8 \\
 & one with $p_\pi{>}0$ & 23{,}764 & 65.0 & 14.4 & 46.1 & 11.3 & 14.6 & 38.9 \\
 & both with $p_\pi{>}0$ & 2{,}818 & 28.0 & 11.0 & 21.6 & 10.5 & 10.9 & 31.5 \\
 & distinct policies & 29{,}743 & 80.0 & 15.3 & 72.3 & 14.5 & 15.1 & 26.9 \\
\addlinespace[2pt]
Garnet $h{\ge}0.5$ & neither with $p_\pi{>}0$ & 85{,}511 & 97.1 & 7.6 & 97.1 & 7.6 & 7.6 & 7.6 \\
 & one with $p_\pi{>}0$ & 53{,}300 & 83.9 & 1.8 & 66.6 & 1.8 & 12.1 & 4.0 \\
 & both with $p_\pi{>}0$ & 8{,}189 & 58.8 & 1.0 & 40.3 & 0.5 & 10.2 & 4.1 \\
 & distinct policies & 70{,}871 & 88.7 & 5.3 & 80.1 & 5.6 & 10.7 & 6.0 \\
\bottomrule
\end{tabular}
}
\end{table}

\paragraph{Pre-specified hypotheses.}
Three hypotheses were fixed in code before the first run; a recorded hash identifies them but does not date them, and
they were not externally registered. Policy values are compared on the normalised score
$z=(J-J_\beta)/(J^*-J_\beta)$, computed per cell, draw and deployment target, where $J^*$ is the target's unconstrained
optimum and $J_\beta$ the behaviour policy's value on the same target: $z=1$ is optimal, $z=0$ matches the behaviour
policy and $z<0$ is worse than it. Statistics use dataset draws as units, 95\% stratified bootstrap intervals
\citep{agarwal2021precipice} and two-sided sign-flip tests with Holm correction \citep{holm1979}.
\begin{itemize}[leftmargin=1.4em, itemsep=1pt, topsep=2pt]
\item \textbf{H1.} In each of the six estimated Gumbel cells, Kendall $\tau_b$ against the truth differs between the mean
and terminate rules. It held in all six cells, always in favour of termination (\cref{tab:h1}).
\item \textbf{H2.} Constraint-aware methods differ in normalised deployment score from projected IQL, an unconstrained
learner made admissible after training (12 treatments, three families, Holm over 36 tests, TOST margin 0.05;
\citealp{schuirmann1987tost}). Over all Gumbel cells pooled, no constraint-aware method beat projected IQL
(\cref{tab:h2}); four were TOST-equivalent and nine were significantly worse. The two groups overlap in masked CQL,
which is significantly worse by a negligible margin and also inside the equivalence margin, so they cover all 12
treatments. In the per-cell scores, which are exploratory and released with the code
(\texttt{results/tables/xenv/xenv\_normalized\_scores.csv}), several methods did beat it with the physician who rarely
gives vasopressors.
\item \textbf{H3.} H3 predicted that, on Garnet, the mean-rule error grows with the size of the harm, $|h|$. The
pre-specified test detected a positive association ($\rho=0.57$, $p=10^{-4}$), but the error actually increases with
$h$ itself (0.010, 0.018, 0.039 and 0.075 at $h=-0.5,0,0.5,1$) and is smallest at $h=-0.5$. The prediction therefore
holds only for $h\ge0$.
\end{itemize}

\begin{table}[htbp]
\centering
\caption{H1: Kendall $\tau_b$ against the truth (default policies, 30 draws). $\Delta=\tau_b(\text{mean})-
\tau_b(\text{terminate})$ with 95\% bootstrap interval and Holm-adjusted sign-flip $p$. Regret: true loss in normalised
score of the benchmark's top choice. The oracle row is exploratory.}
\label{tab:h1}
\resizebox{\linewidth}{!}{\begin{tabular}{lrrrrrr}
\toprule
 & \multicolumn{2}{c}{Kendall $\tau_b$ vs.\ truth} & & & \multicolumn{2}{c}{selection regret ($z$)} \\
\cmidrule(lr){2-3}\cmidrule(lr){6-7}
Gumbel cell & mean rule & terminate & $\Delta$ [95\% CI] & Holm $p$ & mean & terminate \\
\midrule
$n{=}2$k, $\tau{=}5$ & 0.40 & 0.57 & $-$0.18 [$-$0.20, $-$0.15] & $<10^{-4}$ & 0.75 & 0.59 \\
$n{=}2$k, $\tau{=}20$ & 0.49 & 0.93 & $-$0.43 [$-$0.56, $-$0.32] & $<10^{-4}$ & 0.31 & 0.03 \\
$n{=}10$k, $\tau{=}5$ & 0.76 & 0.78 & $-$0.02 [$-$0.04, $-$0.01] & 0.0037 & 0.28 & 0.26 \\
$n{=}10$k, $\tau{=}20$ & 0.70 & 0.83 & $-$0.14 [$-$0.16, $-$0.11] & $<10^{-4}$ & 0.28 & 0.07 \\
$n{=}10$k, $\tau{=}20$, $\epsilon{=}0.2$ & 0.78 & 0.87 & $-$0.10 [$-$0.11, $-$0.09] & $<10^{-4}$ & 0.06 & 0.02 \\
$n{=}10$k, $\tau{=}20$, withhold-vaso & 0.53 & 0.73 & $-$0.20 [$-$0.22, $-$0.19] & $<10^{-4}$ & 0.13 & 0.03 \\
\midrule
oracle construction (exploratory) & 0.80 & 0.95 & $-$0.15 & -- & 0.04 & 0.00 \\
\bottomrule
\end{tabular}
}
\end{table}

\begin{table}[htbp]
\centering
\caption{H2: difference in normalised deployment score between each pre-specified treatment and projected IQL. The
deployment target is the terminate rule on \icu{} and the true system on Gumbel and Garnet. Test units: 50 draws on
\icu{}, 210 (cell, draw) units on Gumbel and 30 draw blocks on Garnet. $^*$: Holm-significant at 0.05 over the 36
two-sided sign-flip tests. IQM: interquartile mean.}
\label{tab:h2}
\small
\begin{tabular}{lrrrrrr}
\toprule
 & \multicolumn{2}{c}{ICU-Sepsis} & \multicolumn{2}{c}{Gumbel} & \multicolumn{2}{c}{Garnet} \\
\cmidrule(lr){2-3}\cmidrule(lr){4-5}\cmidrule(lr){6-7}
Treatment & $\Delta$ mean & $\Delta$ IQM & $\Delta$ mean & $\Delta$ IQM & $\Delta$ mean & $\Delta$ IQM \\
\midrule
BC [masked] & $-$0.013$^{*}$ & $-$0.013 & 0.000 & 0.000 & $-$0.129$^{*}$ & $-$0.157 \\
CQL [masked] & $-$0.023$^{*}$ & $-$0.023 & 0.000$^{*}$ & 0.000 & $-$0.160$^{*}$ & $-$0.185 \\
IQL [masked] & 0.000 & 0.000 & 0.000 & 0.000 & $+$0.001 & $+$0.001 \\
BCQ [masked] & $+$0.023$^{*}$ & $+$0.023 & 0.000 & 0.000 & $+$0.243$^{*}$ & $+$0.083 \\
VI-LCB [masked] & $+$0.050$^{*}$ & $+$0.050 & $-$0.175$^{*}$ & $-$0.218 & $+$0.397$^{*}$ & $+$0.171 \\
MBS-QI [masked] & $+$0.015$^{*}$ & $+$0.014 & $-$0.671$^{*}$ & $-$0.618 & $+$0.366$^{*}$ & $+$0.155 \\
SPIBB [projected] & $+$0.014$^{*}$ & $+$0.015 & $-$0.383$^{*}$ & $-$0.403 & $+$0.009 & $-$0.283 \\
Lagrangian-MB [native] & $+$0.018$^{*}$ & $+$0.018 & $-$0.857$^{*}$ & $-$0.544 & $+$0.062$^{*}$ & $-$0.050 \\
CMDP-LP (model) [native] & $-$0.005 & $-$0.011 & $-$0.796$^{*}$ & $-$0.523 & $-$0.021 & $-$0.195 \\
SfBC (exact) [masked] & $-$0.013$^{*}$ & $-$0.013 & $-$0.198$^{*}$ & $-$0.265 & $+$0.010 & $-$0.294 \\
FISOR-tab (exact) [native] & $-$0.012$^{*}$ & $-$0.013 & $-$0.197$^{*}$ & $-$0.263 & $+$0.012 & $-$0.292 \\
Diffusion (exact, guided) [native] & $+$0.052$^{*}$ & $+$0.052 & $-$0.051$^{*}$ & $-$0.106 & $+$0.402$^{*}$ & $+$0.171 \\
\bottomrule
\end{tabular}

\end{table}

\paragraph{Sensitivity cells.}
\cref{tab:sens} varies one design choice at a time: softer physicians, which leave more than one supported action in
many states; the timeout hazard $\eta$; the rule that removes dead states; and the state abstraction. With true
supported rows, the NHB-dependent claims stay correct in every variant except the Gumbel cells with the argmax
dead-state rule, where 6 of 43,697 are wrong. With estimated rows, the certified error moves with the amount of data
and with how much of the state the abstraction hides.

\begin{table}[htbp]
\centering
\caption{Sensitivity cells (default policies, 30 draws per cell; reference cell in parentheses). one action: share of
decision states whose mask admits a single action (\%). NHB-only wrong: wrong claims / claims certified only with NHB,
on the true supported rows (--: no true-rows target). certified, cert.\ err.: share of pairs certified by the NHB test and
share of those claims the truth reverses or ties, on estimated rows (\%). point err.: share of strict single-rule claims
that are wrong (mean/terminate). Last column: Kendall $\tau_b$ under the mean rule minus that under the terminate rule.
References: the base cell for the physician and timeout rows; the same cells under the \icu{} dead-state rule; and
the aggregated cell with diabetes hidden for the abstractions.}
\label{tab:sens}
\resizebox{\linewidth}{!}{\begin{tabular}{lrrrrrr}
\toprule
& & \multicolumn{1}{c}{true rows} & \multicolumn{4}{c}{estimated rows} \\
\cmidrule(lr){3-3}\cmidrule(lr){4-7}
variant (reference) & one action & NHB-only wrong & certified & cert.\ err. & point err.\ (mean/term) & $\tau_b$(mean) $-$ $\tau_b$(term) \\
\midrule
physician $\epsilon{=}0.3$ & 81 (96) & 0 / 8{,}382 & 76.5 (72.0) & 3.2 (7.5) & 8.9/5.2 & $-$0.09 ($-$0.14) \\
physician $\epsilon{=}0.5$ & 58 (96) & 0 / 8{,}284 & 79.7 (72.0) & 2.5 (7.5) & 7.4/6.7 & $-$0.03 ($-$0.14) \\
\midrule
timeout $\eta{=}0.02$, 10k & 95 (96) & 0 / 7{,}582 & 72.1 (72.0) & 4.4 (7.5) & 11.2/6.9 & $-$0.10 ($-$0.14) \\
timeout $\eta{=}0.1$, 10k & 96 (96) & 0 / 8{,}010 & 72.2 (72.0) & 13.6 (7.5) & 20.0/13.8 & $-$0.13 ($-$0.14) \\
timeout $\eta{=}0.02$, 2k & 93 (95) & 0 / 7{,}652 & 74.8 (72.5) & 16.5 (23.2) & 21.6/16.7 & $-$0.10 ($-$0.18) \\
timeout $\eta{=}0.1$, 2k & 95 (95) & 0 / 7{,}877 & 72.6 (72.5) & 26.5 (23.2) & 32.4/24.4 & $-$0.16 ($-$0.18) \\
\midrule
dead states by argmax, Gumbel (6) & 96 (92) & 6 / 43{,}697 & 69.2 (68.6) & 8.8 (10.6) & 16.1/10.2 & $-$0.17 ($-$0.18) \\
dead states by argmax, Garnet (2) & 30 (27) & 0 / 7{,}041 & 89.2 (89.7) & 4.4 (6.3) & 9.7/4.5 & $-$0.11 ($-$0.10) \\
\midrule
aggregation: glucose hidden, diabetes shown & 79 (91) & -- & 84.8 (72.5) & 2.2 (43.7) & 6.9/5.4 & $-$0.04 ($-$0.30) \\
aggregation: $k$-means 150 (diabetes hidden) & 77 (91) & -- & 85.4 (72.5) & 12.1 (43.7) & 17.0/13.6 & $-$0.07 ($-$0.30) \\
aggregation: $k$-means 50 (diabetes hidden) & 46 (91) & -- & 86.7 (72.5) & 39.6 (43.7) & 41.5/38.0 & $-$0.07 ($-$0.30) \\
\bottomrule
\end{tabular}
}
\end{table}

\paragraph{Exact error decomposition.}
For a policy learned under rule $c$, let $P_0$ be the true rows with the lifted mask and $P_1$ the true rows completed by
$c$. Then
\[
J_{\mathrm{true}}(\pi)-J_{\mathrm{bench}}(\pi)=\big[J_{P_0}(\pi)-J_{P_1}(\pi)\big]+\big[J_{P_1}(\pi)-J_{\mathrm{bench}}(\pi)\big],
\]
a completion (semantics) term, which \cref{thm:bracket} controls, and an estimation term, which vanishes for the
oracle construction (\cref{tab:decomp}).

\begin{table}[htbp]
\centering
\caption{Exact error decomposition into a semantics term and an estimation term. $|\cdot|$: mean absolute value over the
default policies and 30 draws (terms need not sum). sem.: signed mean semantics term; negative values mean the rule
over-values unsupported actions.}
\label{tab:decomp}
\resizebox{\linewidth}{!}{\begin{tabular}{lrrrrrrrr}
\toprule
 & \multicolumn{4}{c}{mean rule} & \multicolumn{4}{c}{terminate rule} \\
\cmidrule(lr){2-5}\cmidrule(lr){6-9}
Cell & $|\mathrm{total}|$ & $|\mathrm{sem.}|$ & sem. & $|\mathrm{est.}|$ & $|\mathrm{total}|$ & $|\mathrm{sem.}|$ & sem. & $|\mathrm{est.}|$ \\
\midrule
Gumbel: $n{=}2$k, $\tau{=}5$ & 0.045 & 0.016 & $-$0.016 & 0.030 & 0.079 & 0.057 & $+$0.057 & 0.027 \\
Gumbel: $n{=}2$k, $\tau{=}20$ & 0.096 & 0.039 & $-$0.039 & 0.057 & 0.129 & 0.081 & $+$0.081 & 0.054 \\
Gumbel: $n{=}10$k, $\tau{=}5$ & 0.023 & 0.005 & $-$0.005 & 0.019 & 0.038 & 0.022 & $+$0.022 & 0.019 \\
Gumbel: $n{=}10$k, $\tau{=}20$ & 0.028 & 0.016 & $-$0.016 & 0.012 & 0.068 & 0.058 & $+$0.058 & 0.011 \\
Gumbel: $n{=}10$k, $\tau{=}20$, $\epsilon{=}0.2$ & 0.048 & 0.031 & $-$0.031 & 0.017 & 0.096 & 0.081 & $+$0.081 & 0.016 \\
Gumbel: $n{=}10$k, $\tau{=}20$, withhold-vaso & 0.025 & 0.011 & $-$0.011 & 0.015 & 0.058 & 0.047 & $+$0.047 & 0.015 \\
Gumbel: oracle & 0.017 & 0.017 & $-$0.017 & 0.000 & 0.060 & 0.060 & $+$0.060 & 0.000 \\
Garnet: $h{=}{-}0.5$, $\epsilon{=}0.05$ & 0.011 & 0.002 & $+$0.002 & 0.010 & 0.040 & 0.033 & $+$0.033 & 0.011 \\
Garnet: $h{=}{-}0.5$, $\epsilon{=}0.3$ & 0.009 & 0.001 & $+$0.001 & 0.009 & 0.033 & 0.029 & $+$0.029 & 0.010 \\
Garnet: $h{=}0.0$, $\epsilon{=}0.05$ & 0.016 & 0.000 & 0.000 & 0.016 & 0.046 & 0.034 & $+$0.034 & 0.016 \\
Garnet: $h{=}0.0$, $\epsilon{=}0.3$ & 0.021 & 0.000 & 0.000 & 0.021 & 0.025 & 0.010 & $+$0.010 & 0.021 \\
Garnet: $h{=}0.5$, $\epsilon{=}0.05$ & 0.036 & 0.016 & $-$0.016 & 0.020 & 0.030 & 0.014 & $+$0.014 & 0.020 \\
Garnet: $h{=}0.5$, $\epsilon{=}0.3$ & 0.042 & 0.010 & $-$0.010 & 0.033 & 0.034 & 0.011 & $+$0.011 & 0.032 \\
Garnet: $h{=}1.0$, $\epsilon{=}0.05$ & 0.041 & 0.030 & $-$0.030 & 0.012 & 0.012 & 0.000 & 0.000 & 0.012 \\
Garnet: $h{=}1.0$, $\epsilon{=}0.3$ & 0.108 & 0.030 & $-$0.030 & 0.078 & 0.068 & 0.005 & $+$0.005 & 0.071 \\
\bottomrule
\end{tabular}
}
\end{table}

\paragraph{Bootstrap widening and confidence sets.}
On seven cells we re-estimated the rows from $B=10$ bootstrap resamples of the hospital episodes and used the 2.5\%
quantile of the terminate value and the 97.5\% quantile of the mean value (\cref{tab:boot}). This is a heuristic: $B$ is
small, and the mask and the policies stay fixed. Joint 95\% $L_1$ confidence sets for the rows and $d_0$
\citep{weissman2003l1} give valid intervals but certify no pair, with or without NHB, even with radii shrunk to a tenth
(\cref{tab:cs}).

\begin{table}[htbp]
\centering
\caption{Point intervals $[\Jterm,J_{\mathrm{mean}}]$ against bootstrap-ensemble intervals (30 draws per cell, default
policies). $n_{p>0}$: policy-draws with $p_\pi>0$; coverage (\%) and mean width among them; share of pairs certified and
share of certified claims the truth reverses or ties (\%).}
\label{tab:boot}
\resizebox{\linewidth}{!}{\begin{tabular}{lrrrrrrrrr}
\toprule
 & & \multicolumn{2}{c}{coverage} & \multicolumn{2}{c}{mean width} & \multicolumn{2}{c}{certified share} & \multicolumn{2}{c}{certified err.} \\
\cmidrule(lr){3-4}\cmidrule(lr){5-6}\cmidrule(lr){7-8}\cmidrule(lr){9-10}
Cell & $n_{p>0}$ & point & boot & point & boot & point & boot & point & boot \\
\midrule
Gumbel 2k, $\tau{=}5$ & 283 & 96.8 & 97.9 & 0.4039 & 0.4481 & 72.5 & 22.7 & 23.2 & 5.4 \\
Gumbel 10k, $\tau{=}20$ & 301 & 91.0 & 95.3 & 0.3748 & 0.3965 & 72.0 & 44.7 & 7.5 & 1.5 \\
Gumbel 10k, $\tau{=}20$, withhold-vaso & 281 & 96.4 & 97.9 & 0.3093 & 0.3335 & 70.8 & 46.7 & 12.7 & 3.3 \\
Garnet $h{=}{-}0.5$, $\epsilon{=}0.05$ & 300 & 25.7 & 47.0 & 0.1514 & 0.1715 & 82.2 & 56.1 & 12.9 & 4.1 \\
Garnet $h{=}0$, $\epsilon{=}0.05$ & 280 & 67.9 & 88.9 & 0.1812 & 0.2036 & 83.2 & 41.0 & 14.5 & 4.9 \\
Garnet $h{=}0.5$, $\epsilon{=}0.05$ & 275 & 91.6 & 94.9 & 0.1641 & 0.1876 & 90.2 & 68.6 & 4.6 & 1.0 \\
Garnet $h{=}1$, $\epsilon{=}0.05$ & 283 & 31.1 & 66.8 & 0.1589 & 0.1773 & 90.3 & 67.6 & 4.0 & $<$0.1 \\
\midrule
pooled Gumbel (3 cells) & 865 & 94.7 & 97.0 & 0.3630 & 0.3929 & 71.7 & 38.0 & 14.5 & 3.0 \\
pooled Garnet (4 cells) & 1,138 & 53.3 & 73.8 & 0.1637 & 0.1847 & 86.5 & 58.3 & 8.8 & 2.1 \\
pooled (7 cells) & 2,003 & 71.2 & 83.8 & 0.2498 & 0.2746 & 80.2 & 49.6 & 11.0 & 2.4 \\
\bottomrule
\end{tabular}
}
\end{table}

\begin{table}[htbp]
\centering
\caption{Confidence-set certificates (30 draws per cell, 50 default policies per draw; \%). Left: 95\% $L_1$ confidence
sets for the rows and $d_0$ with the NHB test. cover.: share of policies whose interval contains the true value;
width: mean interval width as a share of the value range; unrestr.: share of the rows used by the policies whose
radius is unbounded, at scale 1 and at scale 0.1; cert.: pairs certified at radius scales 1, 0.3 and 0.1; only scale 1 carries the 95\% guarantee. Right:
the point interval $[\Jterm,J_{\mathrm{mean}}]$ on the same draws. Here cover.\ counts all 50 default policies,
including zero-width admissible ones, which rarely match the true value exactly under estimation error; it therefore
differs from the coverage in \cref{tab:boot}, which counts only policies with $p_\pi>0$.}
\label{tab:cs}
\resizebox{\linewidth}{!}{\begin{tabular}{lrrrrrrrr}
\toprule
& \multicolumn{5}{c}{confidence set (95\%), NHB test} & \multicolumn{3}{c}{point interval} \\
\cmidrule(lr){2-6}\cmidrule(lr){7-9}
cell & cover. & width & unrestr. & unrestr.\ (0.1) & cert.\ (1/0.3/0.1) & cover. & cert. & err. \\
\midrule
Gumbel 2k, $\tau{=}5$ & 100 & 95 & 99 & 85 & 0/0/0 & 18.3 & 72.5 & 23.2 \\
Gumbel 10k, $\tau{=}20$ & 100 & 94 & 97 & 69 & 0/0/0 & 18.3 & 72.0 & 7.5 \\
Gumbel 10k, withhold-vaso & 100 & 94 & 97 & 66 & 0/0/0 & 18.1 & 70.8 & 12.7 \\
Garnet $h{=}{-}0.5$ & 100 & 100 & 84 & 19 & 0/0/0 & 5.1 & 82.2 & 12.9 \\
Garnet $h{=}0$ & 100 & 100 & 82 & 15 & 0/0/0 & 12.7 & 83.2 & 14.5 \\
Garnet $h{=}0.5$ & 100 & 100 & 84 & 19 & 0/0/0 & 16.8 & 90.2 & 4.6 \\
Garnet $h{=}1$ & 100 & 100 & 87 & 23 & 0/0/0 & 5.9 & 90.3 & 4.0 \\
\midrule
Gumbel, pooled & 100 & 94 & 98 & 73 & 0/0/0 & 18.2 & 71.7 & 14.5 \\
Garnet, pooled & 100 & 100 & 84 & 19 & 0/0/0 & 10.1 & 86.5 & 8.8 \\
all 7 cells & 100 & -- & 90 & 42 & 0/0/0 & 13.6 & 80.2 & 11.0 \\
\bottomrule
\end{tabular}
}
\end{table}

\section{Reproducibility}\label{app:repro}
Code, exported results, and scripts that rebuild every table, figure and number in this paper are available at
\url{https://github.com/AshrafulKabir7/certified-comparisons-offline-rl}; a manifest of SHA-256 hashes verifies the
files. Every reported number comes from one uninterrupted run of our notebook from an empty folder on a single
consumer GPU (49.1 hours). A preliminary multi-session run, made with an earlier version of the notebook, agrees closely
(\cref{tab:repro}); its record and the corrections made before the final run are documented in the code release.

\begin{table}[htbp]
\centering
\caption{Agreement between the final run, the source of every result in this paper, and the preliminary run. Top: the
19 fixed \icu{} policies, trained on a GPU with 10 seeds. Bottom: the simulators, evaluated exactly.}
\label{tab:repro}
\small
\begin{tabular}{lrr}
\toprule
quantity & final run & preliminary run \\
\midrule
\multicolumn{3}{l}{\emph{\icu{}, 19 fixed policies (GPU-trained learners, 10 seeds)}} \\
19 fixed policies: largest $|\Delta J|$ & \multicolumn{2}{c}{0.0017} \\
certified pairs among the 19 fixed policies (NHB-free) & 129 (99) & 131 (99) \\
\multicolumn{3}{l}{\emph{Simulators (exact evaluation)}} \\
Gumbel certified claims, true rows (wrong) & 152{,}237 (0) & 152{,}237 (0) \\
Gumbel certified error, estimated rows & 10.56\% & 10.56\% \\
Garnet certified error, estimated rows & 9.21\% & 9.19\% \\
per cell, largest $|\Delta|$: Gumbel certified share / error / mean $\tau_b$ & \multicolumn{2}{c}{0.0002 / 0.0000 / 0.0002} \\
\quad Garnet certified share / error / mean $\tau_b$ & \multicolumn{2}{c}{0.0007 / 0.0020 / 0.0037} \\
H1 (six Gumbel cells): identical numbers & \multicolumn{2}{c}{54 of 54} \\
bootstrap: certified error, point / bootstrap & 11.00\% / 2.43\% & 10.95\% / 2.41\% \\
\icu{} median breakdown $\kappa^*$ & 0.0890 & 0.0886 \\
\bottomrule
\end{tabular}

\end{table}

\end{document}